\documentclass[runningheads]{llncs}

\usepackage{eccv}
\usepackage[accsupp]{axessibility}
\usepackage{tikz}
\usetikzlibrary{arrows.meta,shapes,calc,matrix,fit,positioning,backgrounds,decorations.markings}
\usepackage{pgfplots}
\usepackage{pgfplotstable}
\pgfplotsset{compat=1.9}
\usepackage{xstring}

\usepgfplotslibrary{external}
\IfBeginWith*{\jobname}{fig/extern/}{\finalcopy}{}

\tikzstyle{every picture}+=[
	remember picture,
	every text node part/.style={align=center},
	every matrix/.append style={ampersand replacement=\&},
]
\tikzstyle{tight} = [inner sep=0pt,outer sep=0pt]
\tikzstyle{node}  = [draw,circle,tight,minimum size=12pt,anchor=center]
\tikzstyle{op}    = [draw,circle,tight]
\tikzstyle{dot}   = [fill,draw,circle,inner sep=1pt,outer sep=0]
\tikzstyle{pt}    = [fill,draw,circle,inner sep=1.5pt,outer sep=.2pt]
\tikzstyle{box}   = [draw,rectangle,inner sep=3pt]
\tikzstyle{high}  = [black!60]
\tikzstyle{group} = [high,box,opacity=.5]
\tikzstyle{dim1}  = [fill opacity=.3,text opacity=1]
\tikzstyle{dim2}  = [fill opacity=.5,text opacity=1]
\tikzstyle{dim3}  = [fill opacity=.7,text opacity=1]
\tikzstyle{rectc} = [tight,transform shape]
\tikzstyle{rect}  = [rectc,anchor=south west]

\tikzset{every mark/.append style={solid}}
\pgfplotsset{%smooth,
	grid=both, width=\columnwidth, try min ticks=5,
	every axis/.append style={font=\small},
	every axis plot/.append style={thick,mark=none,mark size=1.8,tension=0.18},
	legend cell align=left, legend style={fill opacity=0.8},
	xticklabel={\pgfmathprintnumber[assume math mode=true]{\tick}},
	yticklabel={\pgfmathprintnumber[assume math mode=true]{\tick}},
	nodes near coords math/.style={
		nodes near coords={\pgfmathprintnumber[assume math mode=true]{\pgfplotspointmeta}},
	},
}

\pgfplotsset{
	dash/.style={mark=o,dashed,opacity=0.6},
	dott/.style={mark=o,dotted,opacity=0.6},
	nolim/.style={enlargelimits=false},
	plain/.style={every axis plot/.append style={},nolim,grid=none},
}

\pgfdeclarelayer{bg4}
\pgfdeclarelayer{bg3}
\pgfdeclarelayer{bg2}
\pgfdeclarelayer{bg1}
\pgfdeclarelayer{fg1}
\pgfdeclarelayer{fg2}
\pgfdeclarelayer{fg3}
\pgfdeclarelayer{fg4}
\pgfsetlayers{bg4,bg3,bg2,bg1,main,fg1,fg2,fg3,fg4}

\pgfkeys{/tikz/.cd, aspect/.store in=\aspect, aspect=1}
\pgfkeys{/tikz/.cd, depth/.store in=\depth, depth=.5}
\pgfkeys{/tikz/.cd, stepx/.store in=\step, stepx=1}

\tikzstyle{geom} = [line join=bevel,aspect=1,depth=.5,z={(\depth*\aspect,\depth)}]
\tikzstyle{wire} = [geom,draw,thick]

\def\cx[#1,#2,#3]{#1}
\def\cy[#1,#2,#3]{#2}
\def\cz[#1,#2,#3]{#3}
\def\ex[#1,#2,#3]{#1,0,0}
\def\ey[#1,#2,#3]{0,#2,0}
\def\ez[#1,#2,#3]{0,0,#3}

\usepackage[export]{adjustbox}

\newcommand{\Th}[1]{\textsc{#1}}
\newcommand{\mr}[2]{\multirow{#1}{*}{#2}}
\newcommand{\mc}[2]{\multicolumn{#1}{c}{#2}}

\newcommand{\red}[1]{{\textcolor{red}{#1}}}

\newcommand{\citeme}[1]{\red{[XX]}}
\newcommand{\refme}[1]{\red{(XX)}}

\makeatletter
\newcommand*\bdot{\mathpalette\bdot@{.7}}
\newcommand*\bdot@[2]{\mathbin{\vcenter{\hbox{\scalebox{#2}{$\m@th#1\bullet$}}}}}
\makeatother

\makeatletter
\DeclareRobustCommand\onedot{\futurelet\@let@token\@onedot}
\def\@onedot{\ifx\@let@token.\else.\null\fi\xspace}

\def\eg{\emph{e.g}\onedot} 
\def\ie{\emph{i.e}\onedot} 
 
 \def\vs{\emph{vs}\onedot}

\makeatother
\definecolor{ForestGreen}{RGB}{34,139,34}

\definecolor{DarkenedMagenta}{RGB}{225,0,167}

\newcommand{\cls}{\textsc{[cls]}\xspace}

\newcommand{\ours}{\texttt{REGLUE}\xspace}
\newcommand{\redi}{\texttt{ReDi}\xspace}
\newcommand{\repa}{\texttt{REPA}\xspace}
\newcommand{\reg}{\texttt{REG}\xspace}

\definecolor{Orange}{RGB}{255, 125, 0}
\definecolor{Purple}{RGB}{0.58, 0.44, 0.86}
\definecolor{VibrantMagenta}{RGB}{255,0,197}
\definecolor{TableColorAdditional}{rgb}{0.882, 0.945, 0.882}
\definecolor{TableColorExtra}{rgb}{0.906, 0.855, 0.941}
\definecolor{TableColor}{rgb}{0.80, 0.90, 0.93}
\definecolor{TableColorGrey}{rgb}{0.96, 0.96, 0.96}

\definecolor{TableColorGreyText}{rgb}{0.55, 0.55, 0.55}          % soft gray
\definecolor{TableColorExtraText}{rgb}{0.60, 0.45, 0.80}         % pale purple
\definecolor{TableColorAdditionalText}{rgb}{0.45, 0.65, 0.45}    % pale green
\definecolor{TableColorText}{rgb}{0.25, 0.60, 0.70}              % pale cyan
\definecolor{RegRowText}{rgb}{0.78, 0.68, 0.25}                  % pale yellow

\def\addlegendimage{\csname pgfplots@addlegendimage\endcsname}

\newcommand{\supplautoref}[1]{\hyperref[#1]{Suppl.~\autoref{#1}}}
\newcommand{\supplsec}[1]{\hyperref[#1]{Suppl. Sec.~\ref{#1}}}
\newcommand{\reffigures}[1]{\hyperref[#1]{Figures~\ref{#1}}}

\def\rvf{{\mathbf{f}}}

\def\rvh{{\mathbf{h}}}

\def\rvs{{\mathbf{s}}}

\def\rvx{{\mathbf{x}}}

\def\rvz{{\mathbf{z}}}
\def\rvcls{{\mathbf{cls}}}

\def\vs{{\bm{s}}}

\DeclareMathAlphabet{\mathsfit}{\encodingdefault}{\sfdefault}{m}{sl}
\SetMathAlphabet{\mathsfit}{bold}{\encodingdefault}{\sfdefault}{bx}{n}

\definecolor{RegRow}{HTML}{FFF4C1}
\definecolor{RegBase}{HTML}{FFE08A}

\newcommand{\secref}[1]{\hyperref[#1]{Sec.~\ref{#1}}}
\newcommand{\sectionref}[1]{\hyperref[#1]{Section~\ref{#1}}}
\usepackage[inkscapelatex=false]{svg}
\usepackage{colortbl}
\usepackage{multirow}
\usepackage{tcolorbox}
\usepackage{pifont}%
\usepackage{bm}

\usepackage{eccvabbrv}
\definecolor{customblue}{rgb}{0.21,0.49,0.74}
\usepackage[pagebackref,breaklinks,colorlinks,allcolors=customblue]{hyperref}
\usepackage{graphicx}
\usepackage{booktabs}
\usepackage{amssymb}
\usepackage{mathtools}
\usepackage[normalem]{ulem}
\usepackage{cuted}   % Required for the 'strip' environment
\usepackage{caption} % Required for \captionof

\usepgfplotslibrary{groupplots}
\pgfplotscreateplotcyclelist{auxsched}{
  {brown!70!black,  mark=*},          % Constant 1.0
  {teal!70!black,   mark=square*},    % Constant 0.5
  {blue!70!black, mark=triangle*},  % Step (ours)
  {orange!80!black, mark=diamond*}    % Step (aggressive)
}

\usepackage{siunitx}
\usepackage{tikz}
\usetikzlibrary{plotmarks}
\definecolor{Prop}{HTML}{993500}   % deep orange 800 (good vs blue)
\newcommand{\ptick}{%
  \tikz[baseline=-0.6ex]\node[
    inner sep=0.25ex,
    rounded corners=0.65ex,
    draw=Prop, fill=white, line width=0.35pt
  ]{$\checkmark$};}

\DeclareRobustCommand{\ptick}{%
  \tikz[baseline=-0.6ex]\node[
    inner sep=0.25ex, rounded corners=0.65ex,
    draw=TableColor!20!TableColorText, fill=white, line width=0.35pt
  ]{\checkmark};%
}
\usepackage[accsupp]{axessibility}  % Improves PDF readability for those with disabilities.

\usepackage{wrapfig}
\usepackage{orcidlink}

\newcommand{\maketitlesupplementary}{%
  \begingroup
  \renewcommand{\teaser}{}% <- teaser becomes a no-op
  \maketitle
  \endgroup
}

\makeatletter
\newif\ifsupp@toc
\supp@tocfalse
\newcommand{\supptocstart}{\global\supp@toctrue}

\newcommand{\printsuptoc}{%
  \begingroup
  \setcounter{tocdepth}{2}% show sections/subsections
  \supp@tocfalse
  
  \def\authcount##1{}% <-- This removes the stray "1 1"
  
  \@ifpackageloaded{hyperref}{%
    \let\old@contentsline\contentsline
    \def\contentsline##1##2##3##4{%
      \ifsupp@toc \old@contentsline{##1}{##2}{##3}{##4}\fi
    }%
  }{%
    \let\old@contentsline\contentsline
    \def\contentsline##1##2##3{%
      \ifsupp@toc \old@contentsline{##1}{##2}{##3}\fi
    }%
  }%
  \par\noindent{\large\bfseries Table of Contents}\par\vspace{4pt}
  \@starttoc{toc}%
  \endgroup
}
\makeatother

\newcommand{\FigRef}[1]{\hyperref[#1]{\textcolor{customblue}{Figure~\ref*{#1}}}}

\begin{document}

\title{REGLUE Your Latents with Global and Local\\Semantics for Entangled Diffusion}

\titlerunning{REGLUE Your Latents}

\author{Giorgos Petsangourakis\inst{1,2}\orcidlink{0009-0006-4953-8858} \and
Christos Sgouropoulos\inst{1}\orcidlink{0009-0006-0432-0601} \and
Bill Psomas\inst{3}\orcidlink{0000-0001-5381-0312}\and
Theodoros Giannakopoulos\inst{1}\orcidlink{0000-0003-1634-824X}\and
Giorgos Sfikas\inst{2}\orcidlink{0000-0002-7305-2886}\and
Ioannis Kakogeorgiou\inst{1}\orcidlink{0000-0001-5200-2620}}

\authorrunning{G.~Petsangourakis et al.}

\institute{IIT, National Centre for Scientific Research “Demokritos” \and
University of West Attica \and
VRG, FEE, Czech Technical University in Prague}
\maketitle

\begin{abstract}
\vspace{-20pt}

Latent diffusion models (LDMs) achieve state-of-the-art image synthesis, yet their reconstruction-style denoising objective provides only indirect semantic supervision: high-level semantics emerge slowly, requiring longer training and limiting sample quality. Recent works inject semantics from Vision Foundation Models (VFMs) either externally via representation alignment or internally by jointly modeling only a narrow slice of VFM features inside the diffusion process, under-utilizing the rich, nonlinear, multi-layer spatial semantics available. 
We introduce \ours (Representation Entanglement with Global-Local Unified Encoding), a unified latent diffusion framework that jointly models (i) VAE image latents, (ii) compact local (patch-level) VFM semantics, and (iii) a global (image-level) \texttt{[CLS]} token within a single \texttt{SiT} backbone. A lightweight convolutional semantic compressor nonlinearly aggregates multi-layer VFM features into a low-dimensional, spatially structured representation, which is entangled with the VAE latents in the diffusion process. An external alignment loss further regularizes internal representations toward frozen VFM targets. On ImageNet 256×256, \ours consistently improves FID and accelerates convergence over \texttt{SiT-B/2} and \texttt{SiT-XL/2} baselines, as well as over \repa, \redi, and \reg. Extensive experiments show that (a) spatial VFM semantics are crucial, (b) non-linear compression is key to unlocking their full benefit, and (c) global tokens and external alignment act as complementary, lightweight enhancements within our global–local–latent joint modeling framework. The code is available at \href{https://github.com/giorgospets/reglue}{https://github.com/giorgospets/reglue}.
\end{abstract}
\vspace{-16pt}
\section{Introduction}
\label{sec:intro}

Latent Diffusion Models (LDMs) have become a dominant choice for high-quality image synthesis, largely because they shift the generative paradigm to modeling a compact VAE latent space~\cite{rombach2022high}.
However, training such models is \emph{challenging}: under a single denoising objective, the model must concurrently learn high-level semantics (\emph{what} to generate, \eg objects, layout, relations) and low-level visual details (\emph{how} to generate it, \eg fine-grained appearance)~\cite{yao2025vavae}.
The reconstruction-style denoising loss provides semantic supervision only \emph{indirectly}, so semantic structure emerges slowly, limiting image quality and convergence speed~\cite{Yu2025repa}.

To address these challenges, recent works leverages semantic representations from strong, pretrained Vision Foundation Models (VFMs), accelerating convergence and improving image quality~\cite{zheng2025diffusion,Yu2025repa,wu2025representation, kouzelis2025redi,shi2026latent,zheng2026diffusion,chen2026aligning}. \repa~\cite{Yu2025repa} proposes a representation \emph{alignment} objective to distill VFM features as an external teacher to the diffusion model.
\reg~\cite{wu2025representation} and \redi~\cite{kouzelis2025redi} go a step further: they \emph{jointly model} the image latent and the semantic signal inside the diffusion process. Regarding this signal, \reg uses a single \emph{image-level} (global) representation (\ie, \texttt{[CLS]}), while \redi employs \emph{linearly} PCA-projected \emph{patch-level} (local) VFM features. 

Both \reg and \redi expose only a \emph{narrow}, \emph{low-capacity} semantic slice of the VFM to the diffusion model, under-utilizing the rich, nonlinear, multi-layer, and spatial semantics available. \texttt{REG}’s \texttt{[CLS]} offers informative image-level guidance, but is inherently non-spatial; fine-grained semantics are recovered via an external alignment loss as in \texttt{REPA}~\cite{Yu2025repa}, which distills spatial semantics and boosts generative performance. In contrast, \redi explicitly models patch-level semantics but its linear PCA projection restricts the representations to a low-dimensional linear subspace, limiting the richness and non-linear spatial information.

We argue that a critical design choice for effectively leveraging VFM features is to model spatial semantics within the diffusion model, while preserving their nonlinear, multi-layer information via compact learned representations produced by a \emph{semantic compressor}.
Specifically, jointly modeling patch-level features with VAE latents provides spatial guidance critical for capturing fine-grained structure and yields larger gains than either external feature alignment alone (\repa) or jointly modeling a single image-level token (\reg). These signals remain complementary: an alignment loss and a global \texttt{[CLS]} token can be added as orthogonal auxiliary signals, but the primary improvements stem from spatial joint modeling of compressed patch features within the diffusion model (see~\autoref{tab:ablationinit}).

In our work, we first train a lightweight convolutional \emph{semantic compressor} that maps nonlinear, multi-layer VFM features into a compact, spatially structured, semantics-preserving representation. Then, we introduce a unified diffusion modeling approach that \emph{jointly} models: (i) these compact patch-level (local) semantic features, (ii) an image-level (global) representation, and (iii) the VAE image latents. During training, we also apply a feature-alignment auxiliary loss that aligns diffusion internal features with VFM teacher representations, further improving image
synthesis performance. The overview is shown in~\autoref{fig:teaser}.
In summary, our contributions are:
\begin{enumerate}

  \item We introduce \ours (\emph{Representation Entanglement with Global–Local Unified Encoding}), a unified diffusion framework that jointly models image-level (global) and patch-level (local) VFM semantics with VAE latents, significantly boosting generative performance.

   \item We propose a lightweight \emph{semantic compressor} that aggregates multi-layer VFM features and maps them to a compact, semantics-preserving space. This compact representation enables significant gains via patch-level joint modeling with VAE latents,  strongly improving synthesized image quality.
    
    \item We show that, in \ours, (a) patch-level (local) semantics, (b) image-level (global) semantics, and (c) REPA-style representation alignment act synergistically, delivering substantial gains in image quality and training convergence, while keeping the diffusion model’s parameters and inference-time compute essentially unchanged.

    \item On ImageNet~256$\times$256 generation benchmark, \texttt{SiT-XL/2+}\ours reaches the 1M-step performance of \redi and \reg using less than 30\% and 80\% of their iterations, respectively.
\end{enumerate}

\section{Related Work}
\label{sec:related}

\noindent\textbf{Latent-variable generative modeling.}
Latent-variable models like Variational Autoencoders and Diffusion
Denoising Probabilistic models are core building blocks of modern generative pipelines~\cite{rombach2022high}. There are often underappreciated connections across these families~\cite{prince2023understanding,bond2021deep}: diffusion can be viewed as a hierarchical VAE with a fixed latent dimension and a frozen encoder~\cite{luo2022understanding}, and both are trained via variational approximations~\cite{bishop2006pattern}. Normalizing flows~\cite{kingma2018glow,zhai2025normalizing} have likewise been analyzed through their links to diffusion~\cite{zhang2021diffusion,albergobuilding2023,zand2024diffusion}. VAEs are also closely related to probabilistic PCA, replacing the linear projection with a neural decoder~\cite{tipping1999probabilistic,ghojogh2021factor}. Analyzing models under a unified framework has repeatedly yielded progress: Scalable Interpolant Transformers (\texttt{SiT})~\cite{ma2024sit}, alongside DiT~\cite{peebles2023scalable} and Lightning DiT~\cite{yao2025vavae}, an indispensable component of state-of-the-art generative frameworks~\cite{Yu2025repa,yao2025vavae,shi2025SVG,kouzelis2025redi,wu2025representation}. Our work espouses an analogous rationale, focusing on a holistic, joint modeling of tokenizer (VAE latents) and VFM semantics within a single framework.

\vspace{4pt}
\noindent\textbf{Representation alignment with VFM features.}
Latent diffusion typically uses a VAE first stage to obtain compact image latents, which are optimized for reconstruction rather than semantics; this can slow semantic emergence and 
curb
the ability to represent abstract concepts
~\cite{lecun2022path,assran2023self}. 
To mitigate this, works enhance the \emph{first stage} by aligning or reshaping the latent space: VA-VAE~\cite{yao2025vavae} adds a VFM alignment loss, TexTok~\cite{zha2025language} injects text description embeddings, MAETok~\cite{chen2025masked} argues for discriminative (non-variational) latents, and FA-VAE~\cite{medi2025favae} separates low/high frequencies via wavelets. Complementarily, the \emph{second stage} (the denoiser) can be aligned to VFMs: \repa~\cite{Yu2025repa} distills mid-block features to VFM targets and accelerates convergence; DDT~\cite{wang2025ddt} decouples encoder/decoder; REPA-E~\cite{leng2025repae} pursues end-to-end VAE$+$denoiser alignment; and SVG~\cite{shi2025SVG} focuses on few-step generation. Our approach is complementary: instead of exposing a narrow semantic slice or relying solely on external representation alignment, we \emph{jointly model} global and local VFM semantics with VAE latents via a frozen, lightweight semantic compressor, entangling them within a single diffusion backbone and thereby accelerating convergence and improving generative quality.

\vspace{4pt}
\noindent\textbf{Joint feature generative modeling.} A complementary line of work directly \emph{models} foundation features as observed variables, conceptually related to multimodal diffusion that learns joint spaces across modalities (\eg text-image-video-audio) rather than distilling from them. Examples include CoDi~\cite{tang2023any} for any-to-any generation across modalities, and video methods that entangle appearance with motion signals like VideoJam~\cite{chefer2025videojam}. Closer to our setting, \texttt{REG}~\cite{wu2025representation} proposes a \texttt{SiT}-based model that incorporates compressed tokens with the addition of also modeling a VFM \texttt{[CLS]} token;
Representation Diffusion (\redi) \cite{kouzelis2025redi}
takes this approach one step further and %this is deliberate wording
models spatially referenced high-level features on equal grounds to VAE latents by a Diffusion Transformer.
We argue that these prior works model VFM suboptimally,
either by discarding useful spatial cues \cite{wu2025representation}
or working with 
constrained, linear
projections of high-level semantics.

\vspace{4pt}
\noindent\textbf{Representation learning.} Supervised convolutional networks~\cite{he2016deep} and, more recently, Vision Transformers (ViTs)~\cite{dosovitskiy2020image} have established strong baselines for transferable visual features, but the field has largely shifted toward pretraining regimes that reduce or remove manual labels. Early self-supervised work relies on hand-crafted pretext tasks (\eg patch permutation~\cite{noroozi2016unsupervised} or rotation prediction~\cite{gidaris2018unsupervised}), while modern approaches favor contrastive objectives~\cite{chen2020simple, oord2018representation} and self-distillation~\cite{grill2020bootstrap, dino, oquab2024dinov}, yielding highly transferable features at scale. Transformer-era enables masked image modeling (MIM): from BEiT~\cite{bao2021beit} and MAE~\cite{he2022masked} to hybrid variants like iBOT~\cite{zhou2021ibot} and AttMask~\cite{kakogeorgiou2022hide}. In parallel, vision–language pretraining learns joint embeddings from web-scale image–text pairs~\cite{schuhmann2022laion}. CLIP~\cite{clip} popularizes this paradigm by aligning images and captions with a contrastive objective, yielding strong zero-shot recognition~\cite{imagenet}, retrieval~\cite{kordopatis2025ilias, psomas2025instance}, and segmentation~\cite{stojnic2025lposs} performance.  SigLIP~\cite{siglip} refines the recipe by replacing the softmax with independent sigmoid losses, and SigLIP-2~\cite{siglipv2} further improves transfer via stronger data and training strategies.
\section{Method}
\label{sec:method}

\newcommand{\teaserfig}[1]{\includegraphics[trim={0cm 0cm 0cm 0cm},width=1.0\textwidth,valign=c]{#1}}

\newcommand{\teaser}{%
\vspace{-2pt}
\centering

\teaserfig{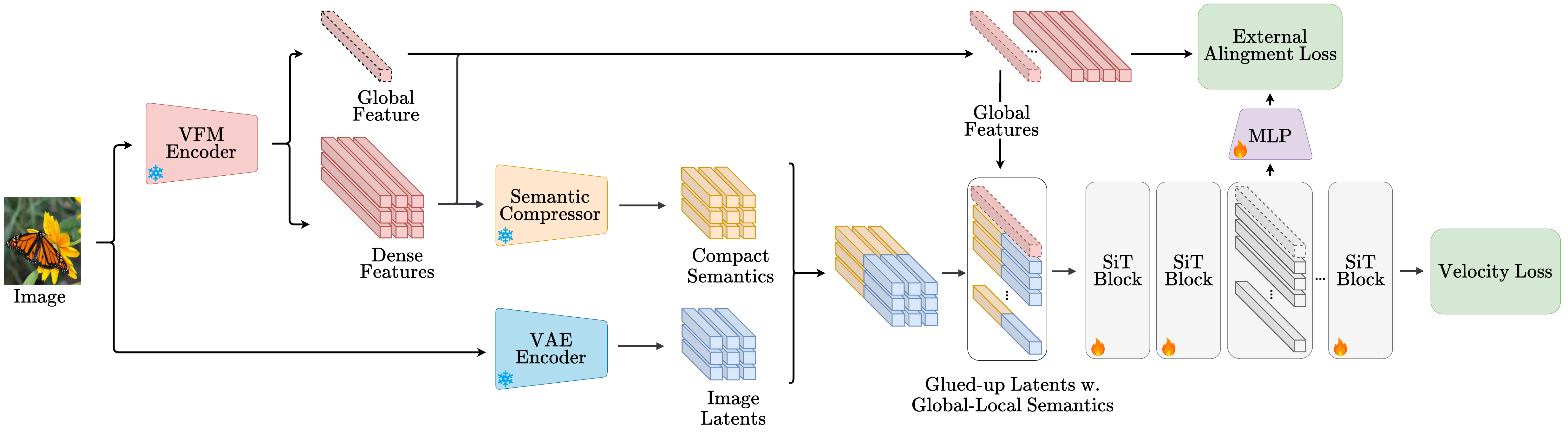}\\

\vspace{-6pt}
\captionof{figure}{\textbf{Overview of} \texttt{REGLUE}, \texttt{R}epresentation \texttt{E}ntanglement with \texttt{G}lobal–\texttt{L}ocal \texttt{U}nified \texttt{E}ncoding. The encoder of a vision foundation model (VFM) provides (i) \emph{local} patch-level features and (ii) a \emph{global} image-level feature (\ie \texttt{[CLS]}). A lightweight semantic compressor (pre-trained offline) maps the patch features to \emph{compact spatial semantics}. In parallel, a frozen VAE encoder produces \emph{image latents}. We concatenate the latents, compressed semantics, and the global token and feed them to a \texttt{SiT} backbone~\cite{ma2024sit}, which \emph{jointly models} all modalities with a velocity objective. Diffusion noise injection is omitted from the illustration. At a selected block, an MLP head applies an external alignment~\cite{Yu2025repa} to match hidden \texttt{SiT} features to clean VFM targets. At sampling, we decode the (jointly) generated VAE latents.}
\label{fig:teaser}

\par\vspace{10pt}
}
\begin{figure*}[t]
\teaser{}
\end{figure*}

\subsection{Preliminaries}
\label{sec:prelim}
\noindent\textbf{Scalable Interpolant Transformer (SiT).} We adopt the \texttt {SiT}~\cite{ma2024sit} framework, built on the stochastic–interpolant formulation~\cite{lipman2023flow,ma2024sit}. Given a clean image $\rvx_\ast$ and a pretrained VAE encoder $\mathcal{E}_z$ that produces image latents $\rvz_\ast \in \mathbb{R}^{D_z \times H_z \times W_z}$, we consider the continuous-time stochastic interpolant:
%----------------------------------------------------
\begin{equation}
\rvz_t \;=\; \alpha_t \rvz_\ast + \sigma_t \bm{\epsilon}_z, 
\qquad \bm{\epsilon}_z \sim \mathcal{N}(\mathbf{0},\mathbf{I}),\ \ t\in[0,1],
\end{equation}
%----------------------------------------------------
where $\alpha_0=\sigma_1=1$ and $\alpha_1=\sigma_0=0$, with $\alpha_t$ decreasing and $\sigma_t$ increasing in $t$. \texttt{SiT} adopts a Transformer-based architecture with \(\mathcal{K}\) stacked blocks, parameterizes the velocity field $\mathbf{v}_\theta(\rvz_t,t)$ and is trained with the standard velocity objective:
%----------------------------------------------------
\begin{equation}
\label{eq:vel-loss}
\mathbb{E}_{\rvz_\ast,\bm{\epsilon}_z,t}\!\left[\;\big\|\mathbf{v}_\theta(\rvz_t,t)\;-\;\dot{\alpha}_t \rvz_\ast\;-\;\dot{\sigma}_t \bm{\epsilon}_z\big\|^2\right].
\end{equation}
Unless stated otherwise, we adopt the linear schedule $\alpha_t = 1 - t$ and $\sigma_t = t$, which yields constant derivatives $\dot{\alpha}_t=-1$ and $\dot{\sigma}_t=1$.

%----------------------------------------------------
\vspace{4pt}
\noindent\textbf{Vision Foundation Model (VFM).} We denote by 
$\mathcal{E}_v(\cdot)$ 
a pretrained VFM (\eg \texttt{DINOv2}), 
which provides \emph{patch-level} semantic features $\rvf_\ast^{(\ell)} \in \mathbb{R}^{D_f \times H_f \times W_f}$ for each layer $\ell \in \{1,2,\ldots,\mathcal{L}\}$, and a global \emph{image-level} representation $\rvcls_\ast \in \mathbb{R}^{D_f}$. $H_f \times W_f$ denotes the patch grid size and $D_f$ the feature dimensionality. 
In ViT-based encoders $H_f,W_f,D_f$ remain fixed across layers.

\iffalse
\subsection{Model overview}

\fi

\subsection{Global--local representation entanglement}
\label{sec:entangle}

Our aim is to \emph{jointly model} (i) VAE latents, (ii) local semantics (patch-level VFM features), and (iii) global semantics (\ie image-level \texttt{[CLS]} token) within a single \texttt{SiT} model. We reuse the notation from~\secref{sec:prelim}. 
Given an image $\rvx_\ast$, $\mathcal{E}_z(\rvx_\ast)=\rvz_\ast\!\in\!\mathbb{R}^{D_z\times H_z\times W_z}$ denotes VAE latents,
$\rvcls_\ast\!\in\!\mathbb{R}^{D_f}$ the global VFM token, 
and
$\rvs_\ast
\!\in\!
\mathbb{R}^{D_s\times H_z\times W_z}$ patch-level \emph{compressed} VFM features 
derived from $\{\rvf_\ast^{(\ell)}\}_{\ell=1}^\mathcal{L}$
and
aligned to the VAE latents' grid
(see \secref{sec:compressor} for a detailed description of the compressor).

\vspace{4pt}
\noindent\textbf{Forward process.}
We first adopt a shared schedule $(\alpha_t,\sigma_t)$ to inject noise for all three 
entangled modalities:
\begin{equation}
\begin{aligned}
\rvz_t   &= \alpha_t \rvz_\ast   + \sigma_t \bm{\epsilon}_z, 
& \bm{\epsilon}_z   &\sim \mathcal{N}(\mathbf{0},\mathbf{I}),\\
\rvs_t   &= \alpha_t \rvs_\ast   + \sigma_t \bm{\epsilon}_s, 
& \bm{\epsilon}_s   &\sim \mathcal{N}(\mathbf{0},\mathbf{I}),\\
\rvcls_t &= \alpha_t \rvcls_\ast + \sigma_t \bm{\epsilon}_{\mathrm{cls}}, 
& \bm{\epsilon}_{\mathrm{cls}} &\sim \mathcal{N}(\mathbf{0},\mathbf{I}),
\end{aligned}
\end{equation}
with independent noise terms and $t\!\in\![0,1]$.

\vspace{4pt}
\noindent\textbf{Velocity objective.}
\texttt{SiT} parameterizes a velocity field $\mathbf{v}_\theta(\rvz_t, \rvs_t, \rvcls_t ,t)$ over the joint state \((\rvz_t,\rvs_t,\rvcls_t)\) and is trained with the multimodal velocity loss:
\begin{equation}
\label{eq:pred}
\begin{aligned}
\mathcal{L}_{\mathrm{v}}
&=  \mathbb{E}\Big[
\;\big\|{\mathbf{v}^{z}_\theta}(\rvz_t, \rvs_t, \rvcls_t ,t) - \dot{\alpha}_t \rvz_\ast - \dot{\sigma}_t \bm{\epsilon}_z\big\|_2^2 \\
&+ \lambda_s \,\big\|{\mathbf{v}^{s}_\theta}(\rvz_t, \rvs_t, \rvcls_t ,t) - \dot{\alpha}_t \rvs_\ast - \dot{\sigma}_t \bm{\epsilon}_s\big\|_2^2 \\
&+ \lambda_{\mathrm{cls}} \,\big\|{\mathbf{v}^{cls}_\theta}(\rvz_t, \rvs_t, \rvcls_t ,t) - \dot{\alpha}_t \rvcls_\ast - \dot{\sigma}_t \bm{\epsilon}_{\mathrm{cls}}\big\|_2^2 \Big] ,
\end{aligned}
\end{equation}

\noindent where $\mathbf{v}^{z}_\theta$, $\mathbf{v}^{s}_\theta$ and $\mathbf{v}^{cls}_\theta$ correspond to the predictions for the VAE latent, global VFM token, and patch-level VFM features velocity; $\lambda_s$ and $\lambda_\mathrm{cls}$ are weighting coefficients. 

\vspace{4pt}
\noindent\textbf{Tokenization and fusion.}
The \texttt{SiT} diffusion backbone operates on a sequence of tokens with a shared width \(D\), so we aim to bring the different modalities 
to a common dimensional space. 
Concretely, we first \emph{patchify}\footnote{Partitioning a tensor \(x\in\mathbb{R}^{C\times H\times W}\) into non-overlapping \(p\times p\) tiles (here \(p{=}2\); denoted \texttt{SiT/2}), flattening each tile, and stacking them row-major into a sequence of patch tokens, resulting in \(x'\in\mathbb{R}^{(HW/p^2)\times(Cp^2)}\).} $\rvz_t$ and $\rvs_t$, forming $\rvz'_t=\mathrm{patch}(\rvz_t)\!\in\!\mathbb{R}^{N\times D_{z'}}$ and $\rvs'_t=\mathrm{patch}(\rvs_t)\!\in\!\mathbb{R}^{N\times D_{s'}}$ (where $N{=}H_zW_z/p^2$ is the number of $p\times p$ patches). 
We then project each 
%stream 
modality 
to the model width \(D\) with linear embedding layers:
\[
\tilde{\rvz}_t=\rvz'_t\mathbf{W}_z,\quad
\tilde{\rvs}_t=\rvs'_t\mathbf{W}_s,\quad
\tilde{\rvcls}_t=
\rvcls_t\mathbf{W}_{\mathrm{cls}}
\]
where $\mathbf{W}_z\!\in\!\mathbb{R}^{D_{z'}\times D}$, $\mathbf{W}_s\!\in\!\mathbb{R}^{D_{s'}\times D}$, and $\mathbf{W}_{\mathrm{cls}}\!\in\!\mathbb{R}^{D_f\times D}$ are learned embedding matrices. 

To combine and jointly model $\tilde{\rvz}_t$ (VAE latents) and $\tilde{\rvs}_t$ (local semantics), there are two straightforward options: (i) concatenate image latents and semantic features along the \emph{sequence} dimension and pass $2N$ patch tokens through \texttt{SiT}, or (ii) \emph{merge channel-wise} and keep a single grid of $N$ tokens. 
We adopt (ii),
avoiding the $2\times$ longer sequence and quadratic self-attention overhead of (i) (see~\supplsec{sec:fusion_strategy}).
The global \cls\ is inherently a single token; thus, we always keep it as a separate token, adding a negligible throughput overhead. 
Finally, the input sequence to the \texttt{SiT} Transformer is:
\begin{equation}
\label{eq:concat}
\rvh^{0}_t
=\Big[
\underbrace{\tilde{\rvcls}_t}_{1\times D}\,;\;
\underbrace{\big(\tilde{\rvz}_t+\tilde{\rvs}_t\big)}_{N\times D}
\Big]
\in\mathbb{R}^{(1+N)\times D},
\end{equation}
where “\([\,;\,]\)” denotes concatenation along token dimension.
%----------------------------------------------------

\vspace{4pt}
\noindent\textbf{Prediction heads.}
Let $\rvh^{\mathcal{K}}_t \in \mathbb{R}^{(1+N)\times D}$ denote the hidden sequence after the last (\(\mathcal{K}\)-th) \texttt{SiT} block. We obtain the per-modality velocity predictions as:
\begin{equation}
\label{eq:heads}
\begin{aligned}
\mathbf{v}^{z}_\theta \;=\; \mathrm{unpatch}\!\big(\,\rvh^{\mathcal{K}}_{t}[1{:}N]\,\mathbf{W}^{z}_{\mathrm{dec}}\,\big),\\
\mathbf{v}^{s}_\theta \;=\; \mathrm{unpatch}\!\big(\,\rvh^{\mathcal{K}}_{t}[1{:}N]\,\mathbf{W}^{s}_{\mathrm{dec}}\,\big),\\
\mathbf{v}^{\mathrm{cls}}_\theta \;=\; \rvh^{\mathcal{K}}_{t}[0]\,\mathbf{W}^{\mathrm{cls}}_{\mathrm{dec}}.
\end{aligned}
\end{equation}
where $\mathbf{W}^{z}_{\mathrm{dec}}\in\mathbb{R}^{D\times D_{z'}},\mathbf{W}^{s}_{\mathrm{dec}}\in\mathbb{R}^{D\times D_{s'}}$ are linear prediction heads, $\mathrm{unpatch}(\cdot)$ reshapes the $N$ patch tokens back to spatial tensors of shape $D_z\times H_z\times W_z$ and $D_s\times H_z\times W_z$, respectively; $\mathbf{W}^{\mathrm{cls}}_{\mathrm{dec}}\in\mathbb{R}^{D\times D_f}$ projects the global token.

%---------------------------------------------------------------------------------
\vspace{4pt}
\noindent\textbf{External representation alignment.} At a selected Transformer block $k\in\{1,\dots,\mathcal{K}\}$, we encourage \texttt{SiT} hidden tokens to stay close to \emph{clean}, frozen VFM targets via a lightweight projector $\phi:\mathbb{R}^{D}\!\to\!\mathbb{R}^{D_f}$ and a cosine loss, following \cite{Yu2025repa}.
Let \(\rvh_t^{k}\in\mathbb{R}^{(1+N)\times D}\) be the hidden sequence at block $k$.
We form the target token sequence by concatenating the global token with patch-level VFM features:

\label{eq:repa-target}
\begin{equation}
\mathbf{y}_\ast \;=\; \big[\rvcls_\ast \,;\, \tilde{\rvf}_\ast^{(\mathcal{L})}\big] \in \mathbb{R}^{(1+N)\times D_f}.
\end{equation}
where $\tilde{\rvf}_\ast^{(\mathcal{L})}$ denotes flattened spatial dimension to tokens.

The representation alignment loss is

\begin{equation}
\label{eq:repa}
\mathcal{L}_{\mathrm{REPA}}
= -\,\mathbb{E}\!\left[\frac{1}{N+1}\sum_{n=1}^{N+1}
\mathrm{sim}\!\left(\mathbf{y}_\ast^{[n]},\; \phi\!\big(\rvh^{k}_t\big)^{[n]}\right)\right],
\end{equation}
where we apply alignment on the \texttt{[CLS]} position and the $N$ semantic patch tokens and $\mathrm{sim}$ is cosine similarity.
%---------------------------------------------------------------------------------

\vspace{4pt}
\noindent\textbf{Total objective.}
We train with the 
multimodal
velocity loss in Eq.~\eqref{eq:pred} and the auxiliary alignment loss in Eq.~\eqref{eq:repa}:
\begin{equation}
\label{eq:total}
\mathcal{L}_{\mathrm{total}} \;=\; \mathcal{L}_{\mathrm{v}} \;+\; \lambda_{\mathrm{rep}}\,\mathcal{L}_{\mathrm{REPA}}.
\end{equation}
%---------------------------------------------------------------------------------

\noindent\textbf{Sampling.} To generate new samples, we employ the reverse-time SDE Euler-Maruyama~\cite{song2020score} using $\mathbf{v}_\theta$
to obtain $(\rvz_0,\rvs_0,\rvcls_0)$ from Gaussian noise, and reconstruct the image via the
(frozen) VAE decoder.

\subsection{A lightweight spatial semantic compressor}
\label{sec:compressor}

Our goal is to jointly model VAE latents with VFM semantics. Naïvely fusing patch-level VFM features with image latents substantially widens the dimensionality to $D_z{+}D_f$ with $D_f\!\gg\!D_z$, biasing \texttt{SiT} capacity toward the representation 
modality 
and hurting generative quality (a phenomenon also addressed in \redi~\cite{kouzelis2025redi} via PCA). 

\vspace{4pt}
\noindent\textbf{Convolutional autoencoder.} Instead of a linear subspace, we introduce a \emph{nonlinear}, lightweight \textit{semantic compressor} $\mathcal{E}_\psi$ that preserves spatial structure while re-balancing dimensionality.
We instantiate $\mathcal{E}_\psi$ as a shallow convolutional autoencoder, pretrain it once to reconstruct VFM features and then keep it frozen. Our compressor finally projects compressed features of dimension $D_s \ll \sum_{\ell} D_\ell$. An overview of the architecture can be found at Suppl. \autoref{fig:compressor_overview}.

\vspace{4pt}
\noindent\textbf{Multi-layer aggregation.}
Since leveraging representations across depths has shown benefits to many scene understanding tasks~\cite{ranftl2021vision,Long_2015_CVPR,lin2017feature,cheng2022masked,zhao2017pyramid,chen2017deeplab}, we typically aggregate the multi-layer patch-level VFM features by channel-wise concatenation:
%----------------------------------------------------

\begin{equation}
\rvf_\ast = \big[\rvf_\ast^{(1)},\rvf_\ast^{(2)},\dots,\rvf_\ast^{({\mathcal{L}})}]\in \mathbb{R}^{\left(\mathcal{L}\cdot D_f\right)\times H_f \times W_f},
\end{equation}
%----------------------------------------------------
where “\([\,,\,]\)” denotes concatenation along the channels.

\vspace{4pt}
\noindent\textbf{Training and inference.}
A lightweight convolutional autoencoder $(\mathcal{E}_\psi,\mathcal{D}_\psi)$ is trained (offline) to reconstruct $\rvf_\ast$:
%----------------------------------------------------
\begin{equation}
\label{eq:ae-train}
\min_{\psi}\; \mathbb{E}\Big[\,\|\mathcal{D}_\psi(\mathcal{E}_\psi(\rvf_\ast)) - \rvf_\ast\|_2^2\,\Big],
\end{equation}
%----------------------------------------------------
We then \emph{freeze} $\mathcal{E}_\psi$, produce compact spatial semantics $\mathcal{E}_\psi(\rvf_\ast)\in\mathbb{R}^{D_s \times H_f \times W_f}$, and spatially resample them to the VAE latent grid (\(H_z\times W_z\)) resulting in $\rvs_\ast\!\in\!\mathbb{R}^{D_s\times H_z\times W_z}$, typically using bilinear resampling (See~\supplsec{sec:resampling}).
\section{Experiments}
\label{sec:experiments}

\subsection{Setup}

\noindent\textbf{Implementation details.}
We strictly follow the standard training protocols of \texttt{SiT} \cite{ma2024sit}. Our experiments are conducted on the ImageNet \cite{deng2009imagenet} dataset. Following the ADM preprocessing pipeline \cite{dhariwal2021adm}, all images are center-cropped and resized to $256 \times 256$ resolution. Each image is then encoded into a latent representation $\rvz_\ast \in \mathbb{R}^{4 \times 32 \times 32}$ 
%$z \in \mathbb{R}^{4 \times 32 \times 32}$
using the pre-trained \texttt{SD-VAE-FT-EMA} \cite{rombach2022high}. Our main experiments are based on \texttt{SiT-B/2} models, which use a $2 \times 2$ patch size and are trained for 400K steps. To assess the impact of our approach at larger scales and longer training, we additionally train \texttt{SiT-XL/2} models for 1M steps. We maintain a batch size of 256 for all experiments.

Unless stated otherwise, for semantic feature extraction, we employ \texttt{DINOv2-B} \cite{darcet2023vision, oquab2024dinov}, concatenating features from blocks 9-12 to obtain a $3072$-channel ($768 \times 4$) feature map at $16 \times 16$ spatial resolution. Our lightweight convolutional autoencoder $(\mathcal{E}_\psi, \mathcal{D}_\psi)$, with a hidden layer of 256, is pretrained for 25 epochs using a reconstruction loss (Eq.~\ref{eq:ae-train}) to compress these maps. The encoder $\mathcal{E}_\psi$ produces a compact 16-channel, $16 \times 16$ latent representation.

For external representation alignment we follow the formulation in ~\secref{sec:method}. We apply the external alignment using  the local and global representations derived from the VFMs last layer, with layer k of the \texttt{SiT} backbone. We use k = 4 for \texttt{SiT-B/2} and k = 8 for \texttt{SiT-XL/2}. Regarding the total objective coefficients we set $\lambda_s=1$, $\lambda_{cls} = 0.03$, and $\lambda_{rep} = 0.5$. More implementation details regarding the semantic compressor and \texttt{SiT}, are provided in~\supplsec{sec:compressor_details} and~\supplsec{sec:sit_details}.

\vspace{4pt}
\noindent\textbf{Evaluation.}
In order to evaluate image generation quality, we report a standard set of quantitative metrics. These include Fréchet Inception Distance (FID) \cite{heusel2017gans} for perceptual quality, sFID \cite{nash2021generating} for spatial coherence, Inception Score (IS) \cite{salimans2016improved} for diversity, as well as Precision (Pre.) and Recall (Rec.) \cite{kynkaanniemi2019improved} to measure sample fidelity and distribution coverage, respectively. All metrics are computed using 50,000 generated samples, following the standard ADM evaluation suite \cite{dhariwal2021adm}. We report additional generation quality metrics in~\supplsec{sec:additional_metrics}. For all experiments, we use Euler–Maruyama SDE sampling with 250 steps. When using Classifier-Free Guidance (CFG)~\cite{ho2022classifier}, we set CFG scale at $w = 2.3$ and guidance interval to $[0, 0.9]$, following~\cite{Kynkaanniemi2024}.

\subsection{Analyzing VFM semantics for generation}

\begin{table}[t]
\small
\centering
\setlength{\tabcolsep}{4.5pt}
\caption{\textbf{Impact of VFM semantics on} \texttt{SiT-B/2} \textbf{for improved generation. Results at 400K training steps.} (a) Baseline \texttt{SiT-B/2} in \textcolor{TableColorGreyText}{\textbf{Gray}}, (b) \repa in \textcolor{TableColorExtraText}{\textbf{Purple}}, (d) \redi in \textcolor{TableColorAdditionalText}{\textbf{Light Green}}, (h) \reg in \textcolor{RegRowText}{\textbf{Yellow}}, and (k-p) \ours (ours) in \textcolor{TableColorText}{\textbf{Light Cyan}}. \texorpdfstring{\ptick}{\checkmark} denotes novel components proposed in our work. While the nonlinear patch-level semantics alone yield substantial gains, the other listed components provide additional improvements. \textsuperscript{†}Setting (p) uses a stronger \texttt{DINOv3-B} VFM; for fairness and consistency with prior work, all other experiments adopt \texttt{DINOv2-B} as the default VFM.}
\vspace{-8pt}
\begin{adjustbox}{max width=0.9\linewidth}
\begin{tabular}{c c|c|c|c|c|c|c} 
\toprule
& \mc{3}{\Th{Local}} &  {\Th{Global}} & \multicolumn{2}{c|}{\Th{External}} & \mr{3}{\Th{FID}} \\
 
&  \mc{3}{\Th{(Patch-level)}} &  {\Th{(Image-level)}} & \multicolumn{2}{c|}{\Th{ Alignment}} \\\cmidrule{1-7}
  
&  {\Th{Non-linear}}   & {\Th{Linear}} & {\Th{Multi-Layer}} & {\texttt{[CLS]}}  & {\Th{Patch-Level}}  & {\texttt{[CLS]}} &    \\ \midrule
\rowcolor{TableColorGrey} (a) & &  &  &  & &  & 33.0\\\midrule

\rowcolor{TableColorExtra} (b) & &  &  &  & \checkmark &  & 24.4  \\
(c) & &  &  & \checkmark &  &  & 25.7  \\ 
\rowcolor{TableColorAdditional} (d) & & \checkmark & &  &  &  & 21.4  \\ 
(e) & & \checkmark &  & & \checkmark & & 18.8  \\
(f) & &  &  &\checkmark & & \checkmark & 33.7  \\
(g) & &  &  &\checkmark & \checkmark &  & 15.5  \\
\rowcolor{RegRow} (h) & &  &  & \checkmark & \checkmark & \checkmark & 15.2  \\ 
(i) & & \checkmark &  &\checkmark & \checkmark & \checkmark & 17.4  \\
(j) & & \checkmark & \checkmark &\checkmark & \checkmark & \checkmark & 23.1  \\ \midrule
\rowcolor{TableColor} (k) & \ptick &  &  & &  &  & 14.3  \\
\rowcolor{TableColor} (l) & \ptick & &  &  & \checkmark & & 14.1  \\
\rowcolor{TableColor} (m) & \ptick & &   & \checkmark& \checkmark & \checkmark &13.7 \\
\rowcolor{TableColor} (n) & \ptick & & \ptick & &  & &13.3 \\ 

\arrayrulecolor{black!30}\cmidrule(lr){1-8}\arrayrulecolor{black}

\rowcolor{TableColor} (o) & \ptick & & \ptick & \checkmark & \checkmark & \checkmark &\bf{12.9} \\ 

\rowcolor{TableColor} (p) & \ptick & & \ptick & \checkmark & \checkmark & \checkmark &\bf{12.3}\textsuperscript{†} \\ 

\bottomrule
\end{tabular}
\end{adjustbox}
\label{tab:ablationinit}
\vspace{-10pt}
\end{table}

We leverage our framework to investigate how diffusion modeling in \texttt{SiT-B/2} benefits from: (i) \emph{which} semantics are modeled (global vs. local), (ii) \emph{how} local semantics are compressed (linear vs. non-linear), and (iii) the \emph{degree to which} an external representation-alignment objective contributes to generation quality. The corresponding design choices and their impact on FID are shown in~\autoref{tab:ablationinit}.

\vspace{4pt}
\noindent\textbf{Local (patch-level) outperform global (image-level) semantics.} We first focus on representations modeled \emph{directly} by the diffusion model, without any external alignment. We find that patch-level semantics clearly outperform global-only signals. Relying solely on the global \texttt{[CLS]} token (setting \textbf{(c)}) attains $25.7$ FID, whereas modeling patch-level features (setting \textbf{(d)}, \redi, linear PCA) improves to $21.4$ FID. Both settings substantially outperform the baseline \texttt{SiT-B/2} backbone ($33.0$ FID, setting \textbf{(a)}), but the gap between \textbf{(c)} and \textbf{(d)} underscores that fine-grained \emph{spatial} semantics are pivotal for improved generative modeling.

\vspace{4pt}
\noindent\textbf{Non-linear compression unlocks local guidance.}
Replacing the linear PCA used in \redi with our lightweight \emph{non-linear} semantic compressor boosts patch-level joint modeling: setting \textbf{(k)} reaches $14.3$ FID without any alignment loss, an absolute $7.1$ FID reduction over \redi (setting \textbf{(d)}). Notably, this also surpasses the state-of-the-art \reg baseline (setting \textbf{(h)}, $15.2$ FID), even though \reg combines global modeling with external alignment. Moreover, enriching local guidance by aggregating multi-layer VFM patch-level features before compression (setting \textbf{(n)}) further reduces FID to $13.3$, \emph{without} using any global token or external alignment. The results indicate that rich spatial semantics are a crucial signal, and non-linear compression is key to unlocking their full benefit. For more analysis about why our \emph{non-linear} compression enhances diffusability, see~\secref{sec:compressor_effectveness} and~\autoref{fig:spectral_profiles}.

\vspace{4pt}
\noindent\textbf{External alignment under local and global modeling.} 
We analyze how \texttt{REPA} behaves when applied to \emph{local} and/or \emph{global} semantics. When the backbone does not jointly model patch tokens, only aligning \emph{local} VFM features already provides a strong boost: the original \texttt{REPA} configuration (setting \textbf{(b)}) improves the default \texttt{SiT-B/2} baseline from $33.0$ \textbf{(a)} to $24.4$ FID. \texttt{REPA} also improves the patch-only setting of \redi \textbf{(d)} to \textbf{(e)} (from $21.4$ to $18.8$ FID). A similar pattern appears when starting from a model that only jointly models the global \texttt{[CLS]} token: adding \emph{local-only} external alignment (setting \textbf{(g)}) reduces FID from $25.7$ \textbf{(c)} to $15.5$, and including the global component in the alignment as well (setting \textbf{(h)}) further improves it to $15.2$. This indicates that local patch alignment is the dominant source of improvement, while global alignment provides a smaller, complementary gain. In contrast, aligning \emph{only} the global information without any local alignment (setting \textbf{(f)}) degrades performance from $25.7$ \textbf{(c)} to $33.7$ FID, suggesting that alignment on global features alone is unstable without spatial anchors. Finally, in our setting, adding \texttt{REPA} on top of non-linear patch-level modeling improves FID from $14.3$ \textbf{(k)} to $14.1$ \textbf{(l)}, showing that once strong spatial semantics are jointly modeled, external alignment acts as a mild but consistent performance complement.

\vspace{4pt}
\noindent\textbf{\ours : Joint local-global-latent modeling.}
Building on these observations, we progressively add the global token and alignment to compressed local modeling. Adding REPA-style alignment on top of \textbf{(k)} yields setting \textbf{(l)} with $14.1$ FID (a modest but consistent gain), indicating that external supervision \emph{complements} joint local modeling. Incorporating the global \texttt{[CLS]} and aligning both local and global signals (setting \textbf{(m)}) further improves to $13.7$ FID. Finally, aggregating multi-layer patch features (from the last four VFM blocks) before compression (setting \textbf{(o)}) forms our final \ours unified setting, achieving $12.9$ FID. However, richer VFM semantics improve latent diffusion only when integrated appropriately. Our non-linear semantic compressor is the key mechanism; without it, simply adding more signals can be counterproductive. For instance, naïvely stacking the existing components (setting \textbf{(i)}) or just using multi-layer PCA (setting \textbf{(j)}) yields significantly worse results (17.4 and 23.12 FID, respectively) compared to \ours at 12.9 FID. To study the dependence of \ours on the underlying VFM, we also examine a stronger VFM, \texttt{DINOv3-B}. As reported in setting \textbf{(p)}, \texttt{DINOv3-B} yields the best result (FID $12.3$), improving over \texttt{DINOv2-B} (FID $12.9$) and indicating that \ours can effectively exploit a more powerful semantic encoder. Nevertheless, to remain consistent with prior work and enable fair comparison, we adopt \texttt{DINOv2-B} as our default VFM in all main experiments. For a more in-depth analysis of our compressor, see~\secref{sec:compressor}.
\vspace{-10pt}
\subsection{Enhancing diffusion models}

\noindent\textbf{Accelerating convergence.}
\begingroup
\parfillskip=0pt\relax
\autoref{tab:acceleration}\textcolor{customblue}{(a)} reports conditional ImageNet $256 \times 256$ results without classifier-free guidance (CFG) with a \texttt{SiT-B/2} backbone. \ours reaches $14.5$ FID at $300$K steps surpassing \texttt{REG} ($15.2$ at $400$K) with \emph{$25$\% fewer} iterations, and further improves to $12.9$ at $400$K. At 400K, \ours reduces FID by $60.9$\% vs. vanilla \texttt{SiT-B/2} (33.0), $47.1$\% vs. \texttt{REPA} (24.4), and $39.7$\% vs. \texttt{ReDi} (21.4). Notably, \ours inference-time throughput and parameters remain unchanged. Moving to a larger \texttt{SiT-XL/2} backbone and more training steps, in~\autoref{tab:fid_comparison} we show conditional ImageNet $256\times256$ results (no CFG).\par
\endgroup

\begin{wraptable}[20]{r}{0.5\textwidth}
\vspace{-30pt}

\captionof{table}{\textbf{Conditional and unconditional generation.} Comparison of \texttt{SiT-B/2} with \texttt{REPA}, \texttt{ReDi}, \texttt{REG}, and \ours on ImageNet $256\times256$ without classifier-free guidance (CFG). We report parameter count, inference-time throughput (img/s), iterations, and FID. Throughput measured on 1$\times$A100 with batch size 64 and 250 sampling steps per image.}
\label{tab:acceleration}

\centering
\scriptsize
\setlength{\tabcolsep}{2.5pt} % Adjust column spacing
\begin{tabular}{lcccc}
\toprule
\Th{Model} & \Th{\#Params} & \Th{Thr.} & \Th{Iter.} & \Th{FID$\downarrow$} \\ 
\midrule 

\mc{5}{(a) \Th{Conditional Generation}}\\

\arrayrulecolor{black!30}\cmidrule(lr){1-5}
\texttt{SiT-B/2} & $130\text{M}$ & $4.1$ & $400\text{K}$ & $33.0$ \\
+ \texttt{REPA}  & $130\text{M}$ & $4.1$ & $400\text{K}$ & $24.4$ \\
+ \texttt{ReDi}  & $130\text{M}$ & $4.1$ & $400\text{K}$ & $21.4$ \\
+ \texttt{REG}  & $132\text{M}$ & $3.8$ & $400\text{K}$ & $15.2$ \\
\rowcolor{TableColor} + \ours (ours)  & $132\text{M}$ & $3.8$ & $300\text{K}$ & $\mathbf{14.5}$ \\
\rowcolor{TableColor} + \ours (ours)  & $132\text{M}$ & $3.8$ & $400\text{K}$ & $\mathbf{12.9}$ \\
\midrule

\mc{5}{(b) \Th{Unconditional Generation}}\\

\arrayrulecolor{black!30}\cmidrule(lr){1-5}\arrayrulecolor{black}
\texttt{SiT-B/2} & $130\text{M}$ & $4.1$ & $400\text{K}$ &  $59.8$ \\
+ \texttt{ReDi}   & $130\text{M}$ & $4.1$ & $400\text{K}$ & $43.6$ \\
+ \texttt{REG}   & $132\text{M}$ & $3.8$ & $400\text{K}$ & $29.7$ \\
\rowcolor{TableColor} + \ours (ours)   & $132\text{M}$ & $3.8$ & $400\text{K}$ & $\mathbf{28.7}$ \\

\bottomrule
\end{tabular}
\end{wraptable}

\noindent At $200$K steps, \ours achieves $4.6$ FID, outperforming \texttt{REG} ($5.0$) and substantially surpassing \texttt{REPA} ($11.1$) and \texttt{ReDi} ($12.5$). Notably, \ours reaches $2.7$ FID at $700$K, matching \texttt{REG}’s $1$M performance ($2.7$) with $30\%$ fewer iterations. At $1$M, \ours sets the best score ($2.5$) vs. \texttt{REG} ($2.7$), \texttt{ReDi} ($5.1$), and \texttt{REPA} ($6.4$). Importantly, \ours improves performance without incurring computational overhead, as measured by throughput (\autoref{tab:acceleration}\textcolor{customblue}{(a)}). See also \supplsec{sec:fid_vs_training_compute} for more details on computational efficiency.

\vspace{4pt}
\noindent\textbf{Unconditional generation.}
We evaluate our method in the unconditional setting and summarize the results in~\autoref{tab:acceleration}\textcolor{customblue}{(b)}. The findings are consistent with the analysis in the previous section. Our \texttt{REGLUE} achieves 52\%, 34.2\%, and 3.4\% improvements over \texttt{SiT-B/2}, \texttt{ReDi}, and \texttt{REG}, respectively, demonstrating the effectiveness of nonlinear compression and joint local–global feature modeling. Remarkably, even in this more challenging unconditional setting, \ours (28.7 FID) substantially outperforms the conditional \texttt{SiT-B/2} baseline (33.0 FID).

\vspace{4pt}
\noindent\textbf{State-of-the-art comparison.} \autoref{tab:sota_comparison} reports quantitative results on ImageNet with classifier-free guidance. \ours improves over \texttt{REG} at matched epochs and closes the gap to longer-trained baselines. At $80$ epochs, \ours lowers FID to $1.59$ vs $1.86$ for \texttt{REG}. Notably, \ours at 80 epochs matches the performance of 160-epoch \texttt{REG}. At $160$ epochs, it further improves to $1.46$ vs~$1.59$. 
Although trained for $5\times$ fewer epochs than the 800-epoch variants (\texttt{REPA}, \texttt{ReDi}, \texttt{REG}), the 160-epoch \ours remains competitive with models that leverage VFM representations and are trained for substantially longer (\texttt{REPA}, FID $1.42$; \texttt{REG}, FID $1.36$). Classifier-free guidance ablations are presented in~Suppl.~\autoref{tab:model_cfg_metrics}. We provide qualitative results in~\secref{sec:visualizations}, along with small-scale dataset and higher-resolution results in~\supplsec{sec:small_scale_hr_benchnmarks}.

\begin{table}[t]
\raisebox{175.6pt}{
\begin{minipage}[t]{0.37\linewidth}

%------------------------------------------------------------------------------
\small
\captionof{table}{\textbf{Conditional generation.} Comparison of \texttt{SiT-XL/2} with \texttt{REPA}, \texttt{ReDi}, \texttt{REG}, and \ours on ImageNet $256\times256$ without classifier-free guidance (CFG) under comparable settings. We report parameter count, iterations, and FID.}

\vspace{-8pt}

\centering
\scriptsize
\setlength{\tabcolsep}{0.5pt}
\begin{tabular}{lccc}
\toprule
\Th{Model} & \Th{\#Params} & \Th{Iter.} & \Th{FID$\downarrow$} \\ 
\midrule

\texttt{SiT-XL/2} & $675\text{M}$ & $7\text{M}$ & $8.3$ \\

\arrayrulecolor{black!30}\cmidrule(lr){1-4}\arrayrulecolor{black}
+ \texttt{REPA}  & $675\text{M}$ & $200\text{K}$ & $11.1$ \\
+ \texttt{ReDi}  & $675\text{M}$ & $200\text{K}$ & $12.5$ \\
+ \texttt{REG}  & $677\text{M}$ & $200\text{K}$ & $5.0$ \\

\rowcolor{TableColor} + \ours (ours)  & $677\text{M}$ & $200\text{K}$ & $\mathbf{4.6}$ \\

\arrayrulecolor{black!30}\cmidrule(lr){1-4}\arrayrulecolor{black}

+ \texttt{REPA}  & $675\text{M}$ & $400\text{K}$ & $7.9$ \\
+ \texttt{ReDi}  & $675\text{M}$ & $400\text{K}$ & $7.5$ \\ 
+ \texttt{REG}  & $677\text{M}$ & $400\text{K}$ & $3.4$ \\
\rowcolor{TableColor} + \ours (ours)  & $677\text{M}$ & $400\text{K}$ & $\mathbf{3.2}$ \\

\arrayrulecolor{black!30}\cmidrule(lr){1-4}\arrayrulecolor{black}

 + \texttt{ReDi}  & $675\text{M}$ & $700\text{K}$ & $5.6$ \\
\rowcolor{TableColor} + \ours
(ours)  & $677\text{M}$ & $700\text{K}$ & $\mathbf{2.7}$ \\

\arrayrulecolor{black!30}\cmidrule(lr){1-4}\arrayrulecolor{black}
+ \texttt{REPA} & $675\text{M}$ & $1\text{M}$ & $6.4$ \\
+ \texttt{ReDi} & $675\text{M}$ & $1\text{M}$ & $5.1$ \\
+ \texttt{REG}  & $677\text{M}$ & $1\text{M}$ & $2.7$ \\
\rowcolor{TableColor} + \ours 
(ours)  & $677\text{M}$ & $1\text{M}$ & $\mathbf{2.5}$ \\

\arrayrulecolor{black}\bottomrule

\end{tabular}

\label{tab:fid_comparison}
%------------------------------------------------------------------------------

\end{minipage}
}
\begin{minipage}[b]{0.63\linewidth}
% ------------------------------------------------------------------------------
\small
\captionof{table}{\textbf{Comparison with state-of-the-art.} Quantitative results on ImageNet $256 \times 256$ with classifier-free guidance (CFG). \texttt{REPA}, \texttt{ReDi}, \texttt{REG} and \ours employ an \texttt{SiT-XL/2} model. $^{*}$Does not use SD-VAE latents. $^{\dagger}$Uses class-balanced evaluation protocol.}
\vspace{-8pt}
\centering
\scriptsize

\setlength{\tabcolsep}{1.5pt} % Adjust column spacing
\begin{tabular}{l c c c c c c}
\toprule
\Th{Model} & \Th{Epochs} & \Th{FID$\downarrow$} & \Th{sFID$\downarrow$} & \Th{IS$\uparrow$} & \Th{Pre.$\uparrow$} & \Th{Rec.$\uparrow$} \\
\arrayrulecolor{black}\midrule

\rowcolor{TableColorGrey}\multicolumn{7}{l}{\emph{Autoregressive Models}} \\

 \texttt{VAR}  & $350$ &  $1.80$ & - & $365.4$  & $0.83$ & $0.57$ \\
 \texttt{MagViTv2}  &$1080$ &  $1.78$ & - & $319.4$  & $0.83$ & $0.57$ \\
 \texttt{MAR}  &$800$ &  $1.55$ & - & $303.7$  & $0.81$ & $0.62$ \\

\arrayrulecolor{black!40}\midrule

\rowcolor{TableColorGrey}\multicolumn{7}{l}{\emph{Latent Diffusion Models}} \\

\texttt{LDM} & $200$ & $3.60$ & -  & $247.7$ & $0.87$ & $0.48$ \\ 
\texttt{U-ViT-H/2} & $240$ & $2.29$ &$ 5.68$  & $263.9$ & $0.82$ & $0.57$ \\ 
\texttt{DiT-XL/2}   & $1400$  &    $2.27$ & $4.60$ & ${278.2}$ & ${0.83}$ & $0.57$  \\
\texttt{MaskDiT} & $1600$ &  $2.28$ & $5.67$ & $276.6$ & $0.80$ & $0.61$ \\ 
\texttt{SD-DiT} & $480$ & $3.23$ & -    & -     & -    & -     \\
\texttt{SiT-XL/2}   & $1400$ &     $2.06$ & ${4.50}$ & $270.3$ & $0.82$ & $0.59$ \\
\texttt{FasterDiT}   & $400$ &     $2.03$ & ${4.63}$ & $264.0$ & $0.81$ & $0.60$ \\
\texttt{MDT}   & $1300$ &     $1.79$ & ${4.57}$ & $283.0$ & $0.81$ & $0.61$ \\

\arrayrulecolor{black!40}\midrule
\rowcolor{TableColorGrey}\multicolumn{7}{l}{\emph{Leveraging VFM's Representations}} \\

\texttt{ReDi} & $800$ & $1.61$ & ${4.66}$ & ${295.1}$ & ${0.78}$ & $0.64$ \\

\texttt{REPA} & ${800}$ & ${1.42}$ & ${{4.70}}$ & ${{305.7}}$ & ${0.80}$ & ${0.65}$ \\

\texttt{REG} & $800$ & $1.36$ & ${4.25}$ & ${299.4}$ & ${0.77}$ & $0.66$ \\

\texttt{RAE}$^{*}$ & $800$ & $1.13$ & ${-}$ & ${262.6}$ & ${ 0.78}$ & $0.67$ \\

\texttt{REPA-E}$^{*\dagger}$ & $800$ & $1.12$ & ${4.09}$ & ${302.9}$ & ${0.79}$ & $0.66$ \\

\arrayrulecolor{black!30}\cmidrule(lr){1-7}\arrayrulecolor{black}
\texttt{REG} & $80$ & $1.86$ & ${4.49}$ & ${321.4}$ & ${0.76}$ & $0.63$ \\
\texttt{REPA-E}$^{*}$ & $80$ & $1.67$ & ${4.12}$ & ${266.3}$ & ${0.80}$ & $0.63$ \\
\rowcolor{TableColor}\ours (ours) & $80$ & $1.59$ & ${4.22}$ & ${282.9}$ & ${0.78}$ & $0.64$ \\

\arrayrulecolor{black!30}\cmidrule(lr){1-7}\arrayrulecolor{black}

\texttt{REG} & $160$ & $1.59$ & ${4.36}$ & ${304.6}$ & ${0.77}$ & $0.65$ \\

\rowcolor{TableColor}\ours (ours) & $160$ & $1.46$ & ${4.23}$ & ${293.9}$ & $0.78$ & $0.65$ \\

\arrayrulecolor{black}\bottomrule
\end{tabular}

\label{tab:sota_comparison}
%------------------------------------------------------------------------------

\end{minipage}
\vspace{-10pt}
\end{table}

\subsection{Semantic compressor impact}

As we highlight in~\secref{sec:compressor}, the channel dimensionality of VFM representations is substantially higher than that of image latents, which can lead to degraded performance when fused naïvely. To mitigate this, \redi~\cite{kouzelis2025redi} employs linear PCA to project the representations into an low-dimensional latent space; however, as we show in~\autoref{tab:ablationinit}, this design choice is suboptimal. In contrast, we show that our non-linear CNN-based semantic compressor (Suppl. \autoref{fig:compressor_overview}) can substantially improve generation quality. In this section, we examine its main design choices: the compression dimensionality (\autoref{fig:compression_channels_abl}), the compressor capacity (\autoref{tab:model_size}), the set of VFM layers used as input (\autoref{tab:vfm_layer}), and quantify their effect on both sample quality and efficiency. We further measure how much semantic information is preserved under compression using downstream probing tasks (\autoref{fig:attentive_probing}).

\begin{table}[t]
\raisebox{85pt}{
\begin{minipage}[t]{0.64\linewidth}

%------------------------------------------------------------------------------
\scriptsize
\centering
\begin{tikzpicture}
\begin{axis}[
    width=0.85\linewidth,
    height=4.8cm,
    xlabel={Top-1 accuracy (\%)},
    ylabel={FID (↓)},
    xmin=25, xmax=87,
    ymin=12, ymax=22,
    grid=both,
    legend style={at={(0.90,0.45)},anchor=south east,font=\scriptsize},
]

% 1) DINOv2 PCA-compressed 8 channels
\addplot+[
    only marks,
    mark=*,
    mark size=4.2pt,          % r_8
    mark options={fill=TableColorAdditionalText, draw=TableColorAdditionalText},
]
coordinates {(30.3,21.4)};
\addlegendentry{PCA 8 channels}

% 2) DINOv2 our compressed 8 channels
\addplot+[
    only marks,
    mark=*,
    mark size=4.2pt,
    mark options={fill=TableColorText, draw=TableColorText},
]
coordinates {(46.45,14.3)};
\addlegendentry{Ours 8 channels}

% 3) DINOv2 our compressed 16 channels
\addplot+[
    only marks,
    mark=*,
    mark size=6.18pt,
    mark options={fill=TableColorText, draw=TableColorText},
]
coordinates {(66.02,13.3)};
\addlegendentry{Ours 16 channels}

\addplot[
    thick,
    TableColorExtraText,
    dashed,
] coordinates {(84.2,12) (84.2,22)};
\addlegendentry{768 channels}

\end{axis}
\end{tikzpicture}

\captionof{figure}{\textbf{Attentive probing accuracy vs.\ generation quality on ImageNet for \texttt{DINOv2} patch-level compression variants.}
 Each point reports top-1 attentive probing accuracy~\cite{psomas2025attention} and FID, bubble size reflects feature dimensionality. Our nonlinear compressors (8/16 channels) achieve lower FID at higher accuracy compared to PCA-based compression, while the dashed line indicates the full 768-channel representation.
}
\label{fig:attentive_probing}
%------------------------------------------------------------------------------

\end{minipage}\hfill
}
\hfill
\begin{minipage}[b]{0.3\linewidth}

%-------------------------------------------------------------------------
\centering
\hspace{-20pt}
\begin{tikzpicture}
\begin{axis}[
  width=4.5cm, height=3.5cm,
  xlabel={Compression Channels},
  ylabel={FID (↓)},
  xmin=3.5, xmax=20.5,
  ymin=14, ymax=18.2,
  xtick={4,8,12,16,20},
  ytick={14,15,16,17,18},
  tick style={black},
  tick label style={/pgf/number format/fixed},
  major tick length=2pt,
  axis line style={black},
  enlargelimits=false
]

\addplot+[  color=teal!70!black,
  mark=*,
  mark options={draw=teal!70!black, fill=teal!50!black}, 
  thick]
  coordinates {(4,18.00) (8,14.70) (12,14.43) (16,14.39) (20,15.29)};

\end{axis}
\end{tikzpicture}
\vspace{-3pt}
\caption{\textbf{Performance vs. compression channels.} Ablation of the final compression channels, in \texttt{DINOv2} last layer's representation, using \texttt{SiT-B/2}  trained for 400K steps without \texttt{REPA} loss.}
\label{fig:compression_channels_abl}
%------------------------------------------------------------------------------

\end{minipage}
\vspace{-20pt}
\end{table}

\vspace{4pt}
\noindent\textbf{Semantic preservation under compression.}
\FigRef{fig:attentive_probing} evaluates how well compressed patch-level features retain VFM semantics via attentive probing accuracy on ImageNet~\cite{psomas2025attention} and how this relates to generative quality (FID). Our non-linear semantic compressor preserves semantics much better than linear PCA: with only 8 channels it achieves substantially higher probing accuracy and lower FID than the PCA-compressed \redi features, and increasing to 16 channels further improves both metrics, approaching the full 768-channel \texttt{DINOv2} baseline. In contrast, the PCA-based compression in \redi yields low probing accuracy and only modest FID gains, indicating that non-linear, spatially structured compression is key to preserving semantic information while improving generation quality. Additional semantic preservation analysis with semantic segmentation experiments on Cityscapes~\cite{Cordts_2016_CVPR} are presented in~\supplsec{sec:semantic_preservation}.

\vspace{3pt}
\noindent\textbf{Semantic compression dimensionality.}
\autoref{fig:compression_channels_abl}, examines how the number of compression channels affects generative performance. Aggressive compression (i.e., 4-channels) removes too much information, leading to degraded FID. Performance improves as we increase the number of channels up to 16, but degrades again at 20 channels. This suggests an optimal intermediate subspace: the compressed features preserve essential information to guide generation, yet are compact enough to stay balanced with the 4-channel image latents and not dominate the model’s capacity. We therefore adopt 16 channels for \ours configuration.

\vspace{3pt}
\noindent\textbf{Semantic compressor design.}
Our aim is to keep the compressor lightweight while preserving essential information. In~\autoref{tab:model_size}, we present an ablation over different hidden-layer widths. Reducing the hidden dimensionality from 256 to 128 channels degrades FID, indicating that overly constrained bottlenecks limit the ability to preserve essential semantic information. Increasing the hidden size from 256 to 512 channels yields no further improvement in FID, while doubling the model size and significantly reducing throughput, making this configuration inefficient. Pushing the capacity further (e.g., 1024 hidden size) leads to unstable compressor training. Moreover, increasing the model depth, via additional convolutional layers or residual blocks, exhibits the same instability. Overall, a shallow compressor with 256 hidden size offers the best balance between stability, efficiency, and generative performance.

\begin{table}[h]
\centering

% ---------- LEFT TABLE ----------
\begin{minipage}[t]{0.59\columnwidth}
\scriptsize
\centering
\setlength{\tabcolsep}{1pt}

\captionof{table}{\textbf{Compression model size comparison.} 
Evaluation of compression encoder hidden layer sizes using the last 4 \texttt{DINOv2} layers and \texttt{SiT-B/2} as the diffusion model, trained for 400K steps. 
Representations are compressed to 16 channels. \texttt{REPA} loss is not applied. Input and Output layers are Conv2D(\textit{in}, \textit{out}) layers.
Each middle block is a Residual Block which comprises two \(3\times3\) convolutional layers (stride=1, padding=1) with Batch Normalization and ReLU activation. Samples in Throughput column are VFM local representations.} 
\label{tab:model_size}

\vspace{-5pt}

\begin{tabular}{cccccc}
\toprule
\Th{Input} & \Th{Middle} & \Th{Output} & \Th{\#Params} & \Th{Throughput} & \Th{FID}$\downarrow$ \\
\midrule
(3072,128) & (128,128) & (128,16)&3.8 &32,304 & 14.2  \\
(3072,256) & (256,256) & (256,16) &8.3& 18,578 & \textbf{13.3}  \\
(3072,512) & (512,512) & (512,16) &18.9 &9,510 & \textbf{13.3}   \\
\bottomrule
\end{tabular}
\end{minipage}
\hfill
% ---------- RIGHT TABLE ----------
\begin{minipage}[t]{0.37\columnwidth}
\scriptsize
\centering

\captionof{table}{\textbf{Multi-layer features effect.} 
We compare the generation performance without using  \texttt{REPA} loss across different sets of \texttt{DINOv2} layers. Results are reported at 400K training steps. In all runs VFM patch tokens are compressed to 16 channels and \texttt{SiT-B/2} is used. The compressor input layer is denoted by CNN In Dim column.}
\label{tab:vfm_layer}

\vspace{-5pt}

\begin{tabular}{ccc}
\toprule
\Th{Dino Layers} & \Th{CNN In Dim} & \Th{FID}$\downarrow$ \\
\midrule
12 & 768 & 14.3 \\
3,6,9,12 & 3072 & 16.9 \\
9,10,11,12 & 3072 & \textbf{13.3} \\
\bottomrule
\end{tabular}

\end{minipage}
\vspace{-10pt}
\end{table}

\vspace{3pt}
\noindent\textbf{Choosing VFM layers for compression.}
In \autoref{tab:vfm_layer}, we study how the choice of VFM layers fed into the semantic compressor affects generation performance. Motivated by \cite{oquab2024dinov}, we compare three configurations of \texttt{DINOv2-B} patch features: (i) using only the last layer (12), (ii) using four intermediate layers (3, 6, 9, 12), and (iii) using the last four layers (9–12). In all cases, the compressed features are mapped to 16 channels and fed into the \texttt{SiT-B/2} diffusion model. Using only the final layer yields 14.3 FID, while including shallow intermediate layers (i.e., 3 $\&$ 6) degrades performance to 16.9 FID, indicating that early-layer features do not provide useful semantic guidance to the generation. In contrast, aggregating the last four layers (9–12) leads to 13.3 FID, suggesting that jointly compressing semantically rich, deeper VFM features provides the most beneficial signal for our framework. Ablations about KL regularization of the compressor can be found in Suppl. \autoref{tab:compressor_losses} and utilization of different VFMs in Suppl. \autoref{tab:different_vfm}.

\newcommand{\legsample}[2]{%
\tikz[baseline=-0.6ex]{
  % tight bbox
  \path[use as bounding box] (0,-0.75ex) rectangle (0.65em,0.75ex);
  \draw[#1, line width=0.5pt] (0,0)--(0.65em,0);
  \draw[#1] (0.325em,0) plot[mark=#2, mark size=1.05pt] coordinates {(0,0)};
}%
}

\begin{figure}[t]
\centering
\small

\begin{tikzpicture}
\begin{axis}[
  width=0.55\linewidth, height=0.35\linewidth, % keep original wrapfigure proportions
  grid=both,
  tick label style={font=\scriptsize},
  label style={font=\scriptsize},
  every axis plot/.append style={line width=0.6pt, mark options={solid}, mark size=0.4pt},
  xmin=0, xmax=63,
  xtick={0,10,20,30,40,50,60},
  xlabel={Zigzag Frequency Index},
  ylabel={Normal. Amplitude},
  ymin=-0.1, ymax=1.,
  xlabel style={yshift=4pt},
  ylabel style={yshift=-4pt},
  log ticks with fixed point,
  clip=false,
]

% --- PCA ---
\addplot[brown!70!black, mark=diamond*] coordinates {
(0,1) (1,0.95) (2,0.5606514) (3,0.4182594) (4,0.3901555) (5,0.7228149) (6,0.5541975) (7,0.3223044)
(8,0.3051217) (9,0.3061559) (10,0.2836712) (11,0.2978971) (12,0.3903039) (13,0.3026345)
(14,0.3997924) (15,0.4004053) (16,0.2939215) (17,0.2642635) (18,0.2733994) (19,0.2587636)
(20,0.2177273) (21,0.2690135) (22,0.1907393) (23,0.262527) (24,0.2632503) (25,0.290824)
(26,0.2664655) (27,0.3795003) (28,0.2661155) (29,0.2716644) (30,0.3227544) (31,0.278661)
(32,0.1985578) (33,0.1783032) (34,0.1914999) (35,0.1738692) (36,0.1657713) (37,0.2140941)
(38,0.2046392) (39,0.2600368) (40,0.2556489) (41,0.2862081) (42,0.2275581) (43,0.231353)
(44,0.2132049) (45,0.1971612) (46,0.1755952) (47,0.2000071) (48,0.1589827) (49,0.1690685)
(50,0.213029) (51,0.1829635) (52,0.2401585) (53,0.2644159) (54,0.2013714) (55,0.2218913)
(56,0.1946254) (57,0.155454) (58,0.1680389) (59,0.256021) (60,0.167014) (61,0.1456414)
(62,0.1741402) (63,0.2050192)
};

% --- Semantic compressor ---
\addplot[teal!70!black, mark=square*] coordinates {
(0,1) (1,0.4321705) (2,0.2153714) (3,0.1388116) (4,0.1212186) (5,0.2930841) (6,0.239629) (7,0.08793933)
(8,0.08763497) (9,0.08451369) (10,0.07697073) (11,0.07675725) (12,0.1028536) (13,0.09524487)
(14,0.1221512) (15,0.07573994) (16,0.05812905) (17,0.05524221) (18,0.0550753) (19,0.06221851)
(20,0.05365521) (21,0.05929083) (22,0.04003479) (23,0.05973122) (24,0.06007684) (25,0.08179253)
(26,0.07119823) (27,0.0701623) (28,0.05332643) (29,0.05802972) (30,0.07718445) (31,0.05443538)
(32,0.04786892) (33,0.0398436) (34,0.04427688) (35,0.03392643) (36,0.03568317) (37,0.04818876)
(38,0.045328) (39,0.046358) (40,0.05690338) (41,0.0651698) (42,0.0430623) (43,0.04759163)
(44,0.05548676) (45,0.04607612) (46,0.04193093) (47,0.0379237) (48,0.02874439) (49,0.03586332)
(50,0.05396756) (51,0.04436095) (52,0.05233365) (53,0.04855864) (54,0.04426481) (55,0.04368267)
(56,0.03617288) (57,0.03505758) (58,0.03699886) (59,0.04142684) (60,0.04116453) (61,0.02356532)
(62,0.0423698) (63,0.04206865)
};

% --- RGB baseline ---
\addplot[black, mark=*] coordinates {
(0,1) (1,0.07846761) (2,0.04118111) (3,0.0194198) (4,0.01853684) (5,0.02576722) (6,0.02169953) (7,0.01226758)
(8,0.01172234) (9,0.01133486) (10,0.007458358) (11,0.008202107) (12,0.008004229) (13,0.007791466)
(14,0.01300326) (15,0.007354889) (16,0.005056921) (17,0.005739981) (18,0.006095686) (19,0.005457125)
(20,0.004830434) (21,0.002870889) (22,0.003554691) (23,0.004117628) (24,0.004077194) (25,0.003938647)
(26,0.003214304) (27,0.006575692) (28,0.005580041) (29,0.002094863) (30,0.002434089) (31,0.003030989)
(32,0.003030794) (33,0.002687216) (34,0.002289537) (35,0.001950326) (36,0.00139388) (37,0.001702781)
(38,0.002032757) (39,0.002183199) (40,0.001898054) (41,0.001712147) (42,0.001544652) (43,0.001318449)
(44,0.001349515) (45,0.001341971) (46,0.001572162) (47,0.001287038) (48,0.001199348) (49,0.0009904939)
(50,0.0009924743) (51,0.001019056) (52,0.0009842777) (53,0.001063821) (54,0.0008690346) (55,0.0007470792)
(56,0.0007581939) (57,0.0007863885) (58,0.0005955129) (59,0.0005782046) (60,0.0006451444) (61,0.0005246844)
(62,0.0004800995) (63,0.0004174658)
};

\node[
  anchor=north east,
  draw=black!40,
  rounded corners=1pt,
  fill=white,
  fill opacity=0.88,
  text opacity=1,
  inner xsep=1pt,
  inner ysep=1pt
] at (rel axis cs:0.995,0.995) {%
\scriptsize
\setlength{\tabcolsep}{2.2pt}
\renewcommand{\arraystretch}{0.9}
\begin{tabular}{@{}l S[table-format=2.1] S[table-format=1.2] S[table-format=2.1]@{}}
  & {Acc$\uparrow$} & {MSE$\downarrow$} & {FID$\downarrow$}\\
  \midrule
  \legsample{brown!70!black}{diamond*}\,\textcolor{brown!70!black}{PCA} & 41.4 & 0.98 & 23.1\\
  \rowcolor{TableColor}
  \legsample{teal!70!black}{square*}\,\textcolor{teal!70!black}{Sem.Comp.} & 66.0 & 0.68 & 12.9\\
\multicolumn{4}{@{}l@{}}{\legsample{black}{*}\,\textcolor{black}{RGB (frequency-prof. baseline)}}\\
\end{tabular}
};

\end{axis}
\end{tikzpicture}

\caption{\textbf{Diffusability of Semantic Compressor \vs PCA.} Frequency-profile comparison~\cite{skorokhodov2025improving} of compression strategies, with RGB as reference. The inset shows that our compressor preserves more semantic signal (Acc.,MSE) while yielding a more diffusion-friendly representation by suppressing excess high-frequency components, leading to better generation quality (FID). Both map \texttt{DINOv2} last-4-layer channels to 16-D.}
\label{fig:spectral_profiles}
\end{figure}
\vspace{-1pt}
\noindent\textbf{Semantic compressor effectiveness.}
\label{sec:compressor_effectveness}
In \autoref{fig:spectral_profiles}, we provide a \emph{diffusability} diagnostic. Following~\cite{skorokhodov2025improving}, which identifies inordinate high-frequency components in latents as harmful for diffusion modeling, we compute DCT \emph{frequency profiles} and observe that our compressor suppresses high-frequency components that diffusion models tend to mis-model. PCA exhibits a pronounced high-frequency tail, which is harder to model, whereas our semantic compressor strongly attenuates this tail and yields a spectrum closer to the RGB reference, indicating a more diffusion-friendly representation. Crucially, this does not sacrifice semantics (higher ImageNet probing \Th{Acc.}, lower MSE). Collectively, these improvements enable better generative modeling (lower FID).
\vspace{-10pt}
\section{Conclusion}
\label{sec:conclusion}
\vspace{-3pt}
We introduced \ours, a unified generative model for latent diffusion that
enables efficient entanglement of reconstruction-optimized and semantics-optimized image representations.
By \emph{jointly}
modeling VAE latents with VFM patch-level \& global semantics,
coupled with \emph{lightweight compression and aggregation} components,
we have shown that \ours improves generation FID and accelerate convergence on ImageNet baselines by significant margins.

% \clearpage  % TODO FINAL: This \clearpage needs to be removed from both review and camera-ready versions.

\paragraph{Acknowledgements} 

We thank our colleague Theodoros Kouzelis for fruitful discussions. The project has received funding from the European High-Performance computing Joint Undertaking (JU) under grant agreement No 101234269 and the Greek Ministry of Digital Governance. Bill was supported by the EU Horizon Europe programme MSCA PF RAVIOLI (No. 101205297). AWS resources were provided by the National Infrastructures for Research and Technology GRNET and funded by the EU Recovery and Resiliency Facility. The publication/registration fees were partially covered by the University of West Attica. Also, we acknowledge the EuroHPC Joint Undertaking for awarding this project access to the EuroHPC supercomputer LEONARDO, hosted by CINECA (Italy) and the LEONARDO consortium through an EuroHPC Development Access call (Projects ID EHPC-DEV-2026D02-285 \& EHPC-DEV-2026D03-137).
%Please insert your acknowledgments here.

\bibliographystyle{splncs04}
\bibliography{main}
\clearpage
\newcommand{\setpapermetadata}{%
  \title{REGLUE Your Latents with Global and Local\\Semantics for Entangled Diffusion}
  \titlerunning{REGLUE Your Latents}

  \author{Giorgos Petsangourakis\inst{1,2}\orcidlink{0009-0006-4953-8858} \and
  Christos Sgouropoulos\inst{1}\orcidlink{0009-0006-0432-0601} \and
  Bill Psomas\inst{3}\orcidlink{0000-0001-5381-0312} \and
  Theodoros Giannakopoulos\inst{1}\orcidlink{0000-0003-1634-824X} \and
  Giorgos Sfikas\inst{2}\orcidlink{0000-0002-7305-2886} \and
  Ioannis Kakogeorgiou\inst{1}\orcidlink{0000-0001-5200-2620}}

  \authorrunning{G.~Petsangourakis et al.}

  \institute{IIT, National Centre for Scientific Research ``Demokritos'' \and
  University of West Attica \and
  VRG, FEE, Czech Technical University in Prague}
}

\setcounter{page}{1}

\setpapermetadata
\subtitle{Supplementary Material\vspace{-15pt}}
\maketitlesupplementary

% Marker AFTER title: excludes main paper + both maketitles from the ToC
\addtocontents{toc}{\protect\supptocstart}

{
  \hypersetup{linkcolor=black}
  \printsuptoc
}

\setcounter{figure}{3}
\setcounter{table}{7}

\setcounter{section}{0}
\renewcommand{\thesection}{\Alph{section}}
\renewcommand{\theHsection}{\Alph{section}}

\section{Additional Experimental Results}

%--------------------------------------------------------------------

\subsection{Semantic preservation under compression}

\begin{figure}
    \centering
    \includegraphics[trim={0cm 0cm 0cm 0cm},width=0.85\linewidth]{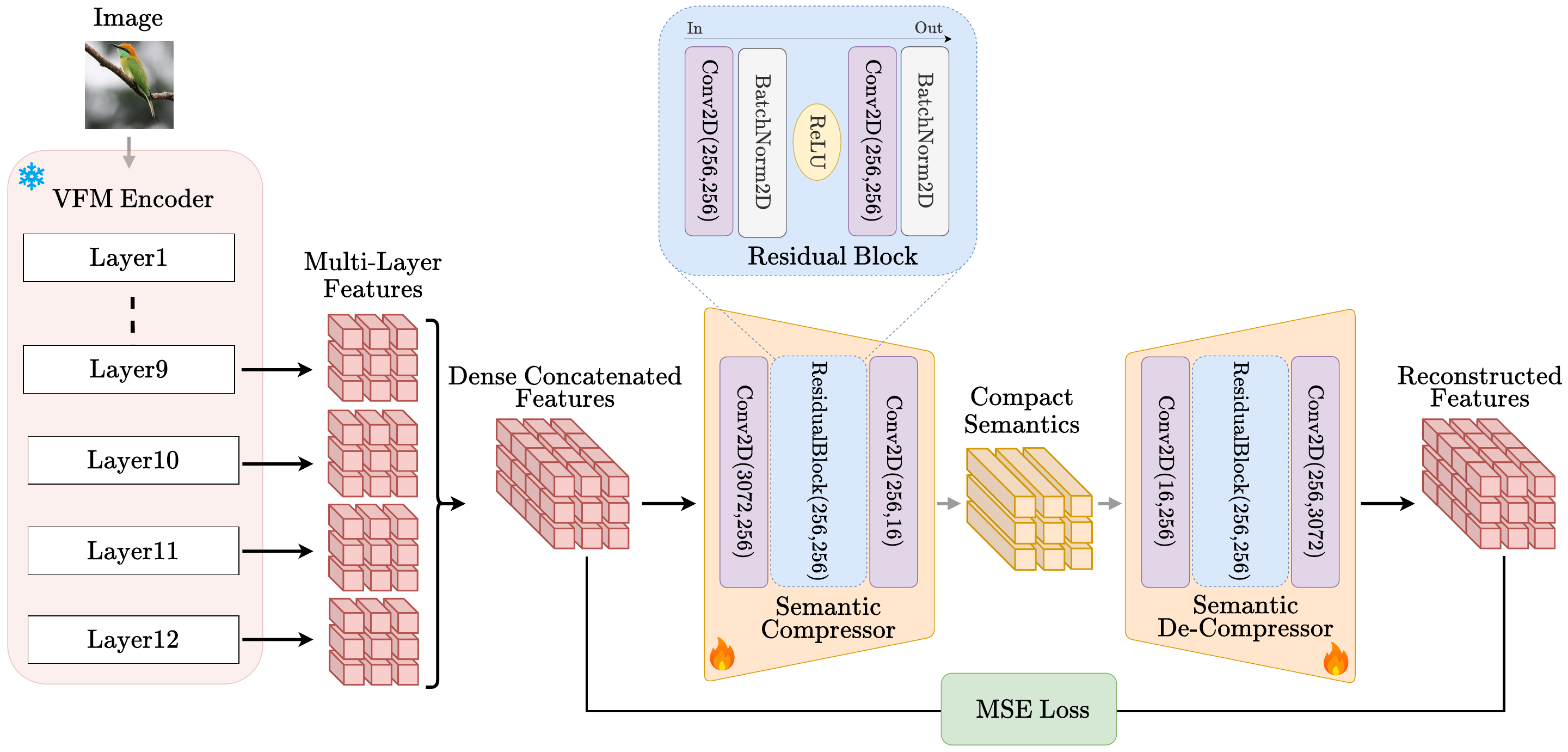}
    \vspace{-4pt}
    % We use \captionof because we are not inside a figure environment
    \captionof{figure}{
    \textbf{Semantic compressor architecture and training.}
    The representations from the last four layers of the vision foundation model (VFM) encoder are concatenated and passed to the compression model, which projects them into a compact 16-channel semantic representation. In our default configuration (corresponding to the middle row of~\autoref{tab:model_size}), the compressor maps the dense concatenated VFM features through an input layer Conv2D(3072, 256), a middle ResidualBlock(256, 256), and an output layer Conv2D(256, 16), where 256 is the hidden dimensionality. The semantic de-compressor then reconstructs the compact semantics back to their original dimensionality. The model is trained using an MSE loss between the dense concatenated features and their reconstructed counterparts.}
    \label{fig:compressor_overview}
\end{figure}

\label{sec:semantic_preservation}

\begin{figure}[t]
\centering
\begin{tikzpicture}
\begin{axis}[
    width=0.75\linewidth,
    height=4.8cm,
    xlabel={mIoU},
    ylabel={FID (↓)},
    xmin=58.5, xmax=73,
    ymin=12, ymax=22,
    grid=both,
    legend style={at={(0.92,0.45)},anchor=south east,font=\footnotesize},
]

% 1) DINOv2 PCA-compressed 8 channels
\addplot+[
    only marks,
    mark=*,
    mark size=4.2pt,          % r_8
    mark options={fill=TableColorAdditionalText, draw=TableColorAdditionalText},
]
coordinates {(59.14,21.4)};
\addlegendentry{PCA 8 channels}

% 2) DINOv2 our compressed 8 channels
\addplot+[
    only marks,
    mark=*,
    mark size=4.2pt,          % same dim → same area
    mark options={fill=TableColorText, draw=TableColorText},
]
coordinates {(67.08,14.3)};
\addlegendentry{Ours 8 channels}

% 3) DINOv2 our compressed 16 channels
\addplot+[
    only marks,
    mark=*,
    mark size=6.18pt,        % ≈ 3 * sqrt(2) so area ∝ 16
    mark options={fill=TableColorText, draw=TableColorText},
]
coordinates {(68.66,13.3)};
\addlegendentry{Ours 16 channels}

\addplot[
    thick,
    TableColorExtraText,
    dashed,
] coordinates {(72.50,12) (72.50,22)};
\addlegendentry{768 channels}

\end{axis}
\end{tikzpicture}
\vspace{-8pt}
\caption{\textbf{Semantic segmentation performance mIoU vs generation quality for different \texttt{DINOv2} patch-level compression variants.} Each point shows the segmentation mIoU on Cityscapes~\cite{Cordts_2016_CVPR} using a DPT~\cite{ranftl2021vision} head on frozen features following implementation from \cite{karypidis2025dinoforesight, yang_depth,yang2024depth} and the FID on ImageNet of the corresponding \texttt{SiT} model. Bubble area is proportional to feature dimensionality. Our non-linear semantic compressors (8 and 16 channels) achieve substantially better FID at higher mIoU than the PCA-compressed features of \texttt{ReDi}. The vertical dashed line indicates the mIoU of the full 768-channel \texttt{DINOv2} representation.}
\label{fig:dpt_segm}
\end{figure}
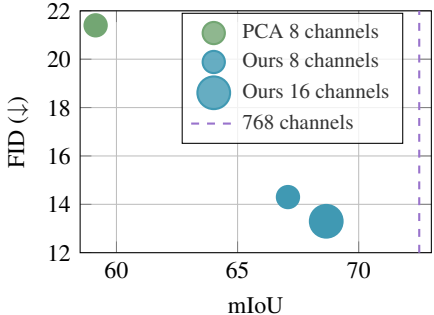

To assess how well our compressed patch-level features preserve vision foundation model (VFM) semantics, we perform an additional experiment on \emph{semantic segmentation}. \autoref{fig:dpt_segm} shows semantic segmentation on Cityscapes~\cite{Cordts_2016_CVPR} measured by mIoU (using DPT~\cite{ranftl2021vision} head on frozen features) versus ImageNet FID for different \texttt{DINOv2} patch–level compression variants. At the same 8-channel compression, our nonlinear compressor achieves 67.1 mIoU{$/$}14.3 FID, notably better than 59.1{$/$}21.4 of PCA ($+$8.0\ mIoU{$/$}$-$7.1 FID). Increasing to 16 channels further improves to 68.7 mIoU{$/$}13.3 FID.
Despite having 96{$\times$} or 48{$\times$} less channels (respectively) than the original 768-channel \texttt{DINOv2} representation (vertical dashed line at 72.5 mIoU), our compressed variants effectively retain most of the semantics while substantially improving generative fidelity, indicating that the learned nonlinear compressor preserves semantic representations and is a better fit than linear PCA in joint semantics-VAE latents modeling.

%--------------------------------------------------------------------
\begin{table}[h]
\footnotesize
\setlength{\tabcolsep}{2pt}
\centering
\caption{\textbf{Semantic compressor loss variants.} ImageNet 256{$\times$}256 comparison without classifier-free guidance (CFG) using \texttt{SiT-B/2}. We compare the impact of different auxiliary objectives in our semantic compressor. We use \texttt{DINOv2-B} last layer representations compressed to 16 channels. For \ours, we follow setting~(k) in~\autoref{tab:ablationinit} (main paper).}
\label{tab:compressor_losses}

\vspace{-8pt}

\begin{tabular}{ccccccc}
\toprule

\Th{Training Objectives}   & \Th{$w_{\mathrm{KL}}$} & \Th{FID}$\downarrow$ & \Th{sFID}$\downarrow$&\Th{Precision$\uparrow$}& \Th{Recall$\uparrow$}  & \Th{Probing Acc.$\uparrow$} \\
\midrule
MSE  & - &\textbf{14.3} & 6.7& 0.62& 0.65  & \textbf{66.01}\\
MSE+KL & $10^{-6}$ &17.2 & 7.1 &0.63 &0.64 & 64.34\\
MSE+KL & $10^{-5}$ & 17.8 & 7.1 &0.63 &0.64 & 63.35\\
MSE+KL & $10^{-3}$ &36.6 & 7.1 &0.63 &0.64 & 44.94\\
\bottomrule
\end{tabular}
\end{table}

%--------------------------------------------------------------------

\subsection{Semantic compressor regularization}

We further ablate the compressor's KL weight. Starting from our MSE-only autoencoder, we add a variational KL term and sweep its weight ($w_{\mathrm{KL}}$), with all other hyperparameters identical to the MSE baseline. We compress the \emph{last} VFM layer to 16 channels and, during diffusion training, model only local semantics without external alignment (similar to setting~(k) in~\autoref{tab:ablationinit}). As shown in \autoref{tab:compressor_losses}, increasing ($w_{\mathrm{KL}}$) consistently degrades both gFID and linear probing accuracy on ImageNet, the latter measuring how well the compressed latent preserves semantic content, while precision and recall stay flat. Plain MSE performs best, and the KL regularization actively harms the semantics that downstream diffusion relies on.

\begin{table}[h]
\footnotesize
\centering
\caption{\textbf{\texttt{REGLUE} with different VFMs.} ImageNet 256{$\times$}256 comparison without CFG using \texttt{SiT-B/2}. For \ours, we follow setting~(o/p) in~\autoref{tab:ablationinit} (main paper).}
\label{tab:different_vfm}

\vspace{-8pt}

\begin{tabular}{ccccc}
\toprule

\Th{VFM} & \Th{FID}$\downarrow$ & \Th{sFID}&\Th{Precision$\uparrow$}& \Th{Recall$\uparrow$}\\
\midrule
\texttt{DINOv2-B} & 12.9 & 5.8& 0.67& 0.63\\
\texttt{DINOv3-B} & \textbf{12.3} & 5.8 &0.67 &0.63\\
CLIP-L  &18.1 & 7.1 & 0.63 & 0.62\\

\bottomrule
\end{tabular}
\vspace{-5pt}
\end{table}

\subsection{Impact of VFM}

To evaluate \ours across different VFMs, we experiment with three encoders: \texttt{DINOv2-B}, \texttt{DINOv3-B}, and CLIP-L. For each backbone, we concatenate the last four layers and adapt the compressor’s input projection to the corresponding embedding size (\eg, $4{\times}768{=}3072$ for \texttt{DINOv2-B}, $4{\times}1024{=}4096$ for CLIP-L). All compressors are trained for 25 epochs with a target compression of 16 channels, and the downstream \texttt{SiT-B/2} generator is trained for 400K steps in every setting for fair comparison. ~\autoref{tab:different_vfm} reports FID, sFID, precision, and recall. \texttt{DINOv3-B} delivers the best generation quality (lowest FID), \texttt{DINOv2-B} is a close second, while CLIP-L lags behind. As already discussed in the main paper, to remain consistent with prior work~\cite{kouzelis2025redi, Yu2025repa, wu2025representation}, we adopt \texttt{DINOv2-B} as our default VFM.

\subsection{Detailed benchmark}
\input{sec/fig_speedup}
We provide a detailed evaluation of \texttt{SiT-XL/2+}\ours with more training iterations and additional metrics.~\autoref{tab:details_reglue} demonstrates the performance, reporting FID, sFID, inception score, precision, and recall. Notably, \ours reaches 7.8 FID at 100K steps, already surpassing the vanilla \texttt{SiT-XL/2} baseline at 7M steps (8.3 FID). It continues to improve substantially, reaching 3.2 at 400K, 2.6 at 750K, and 2.5 at 1M steps. In~\autoref{fig:speedup}, we also show visual examples demonstrating that \ours achieves high-fidelity generations early in training.

\begin{table}[!h]
\centering
\footnotesize
\caption{\textbf{Detailed evaluation} for \texttt{SiT-XL/2+}\ours. ImageNet $256{\times}256$ without CFG.}
\vspace{-8pt}
\setlength{\tabcolsep}{2pt}
\begin{tabular}{lrrrrrrr}

\toprule
\Th{Model} & \Th{\#Iters.} &  \Th{FID$\downarrow$} & \Th{sFID$\downarrow$} & \Th{IS$\uparrow$} & \Th{Prec.$\uparrow$} & \Th{Rec.$\uparrow$} \\
\midrule

\rowcolor{TableColorGrey}  \texttt{SiT-XL/2} 
\ & $7\text{M}$ & $8.3$ & $6.3$  & $131.7$ & $0.68$ & $0.67$ \\

w/ \ours &  $50\text{K}$  & $20.0$ & $6.3$ & $64.7$ & $0.65$ & $0.58$ \\
w/ \ours &  $100\text{K}$ & $7.8$ & $ 4.7$  & $116.0$ & $0.64$ & $0.57$ \\
w/ \ours &  $200\text{K}$ & $4.6$ & $4.4$  & $148.0$ & $0.74$ & $0.63$ \\
w/ \ours &  $400\text{K}$ & $3.2$ & $4.3$  & $171.6$ & $0.75$ & $0.63$ \\
w/ \ours &  $700\text{K}$ & $2.7$ & $4.2$  & $185.0$ & $0.76$ & $0.65$ \\
w/ \ours &  $750\text{K}$ & $2.6$ & $4.1$  & $185.6$ & $0.76$ & $0.65$ \\
w/ \ours &  $1\text{M}$ & $2.5$ & $4.1$  & $188.6$ & $0.76$ & $0.65$ \\

\bottomrule
\end{tabular}
\label{tab:details_reglue}
\end{table}

\subsection{Additional Metrics}
\label{sec:additional_metrics}
~\autoref{tab:additional_metrics} reports CMMD~\cite{Jayasumana_2024_CVPR}, FD and FD$_\infty$~\cite{stein2023exposing}, which provide alternative, robust, and reliable metrics for assessing image quality. At Base scale,
%\gsfikas{(Table 2)}
\ours outperforms \redi and \reg \emph{across all metrics}, indicating the gains are not an artifact of Inception features. At XL, \ours at 1M iters matches or surpasses \reg at 2.4M iters on, surpassing its quality with only 42\% of the training iterations.

\begin{table}[!h]
\centering
\footnotesize
\caption{CMMD, FD, FD$_\infty$ with DINOv2 on ImageNet 256$\times$256. w/o CFG.
}
\vspace{-8pt}
\setlength{\tabcolsep}{2pt}
\begin{tabular}{lcccccc}
\toprule
\Th{Model} & \Th{Iter} & \Th{CMMD$\downarrow$} & \Th{FD$\downarrow$} & \Th{FD$_\infty\downarrow$} & \Th{Precision$\uparrow$} & \Th{Recall$\uparrow$} \\
\midrule
\texttt{ReDi-B}           & 400K & 1.099 & 517.20 & 508.85 & 0.61 & 0.62 \\
\texttt{REG-B}            & 400K & 1.042 & 484.57 & 476.90 & 0.64 & 0.63     \\
\rowcolor{teal!35}
\texttt{REGLUE-B} & 400K & \textbf{0.823} & \textbf{329.89} & \textbf{321.46} & \textbf{0.67} & \textbf{0.64} \\
\midrule
\texttt{SiT-XL}           & 7M   & 0.698 & 233.02 & 225.10 & 0.68 & \textbf{0.67} \\
\texttt{REG-XL}           & 2.4M & 0.479 & 87.48  & 78.65  & \textbf{0.76} & 0.66 \\
\rowcolor{teal!35}
\texttt{REGLUE-XL} & 1M & \textbf{0.458} & \textbf{85.96} & \textbf{77.59} & \textbf{0.76} & 0.65 \\
\bottomrule
\end{tabular}

\label{tab:additional_metrics}
\end{table}

\subsection{Classifier-free guidance}

We provide more evaluation results for classifier-free guidance scales and guidance intervals. We denote by $w$ the CFG scale applied to the VAE latents and the VFM representations, and use VAE-Only to refer to the setting where CFG is applied exclusively to the VAE latents. We also vary the guidance \emph{interval} $[0, \tau]$, following~\cite{Kynkaanniemi2024}.~\autoref{tab:model_cfg_metrics} presents ImageNet 256{$\times$}256 results for \texttt{SiT-XL/2}+\texttt{REGLUE} at 800K steps.

\begin{table}[t]
    \centering
    \footnotesize
    \setlength{\tabcolsep}{2.1pt}
    \caption{\textbf{CFG ablations on \texttt{SiT-XL/2}+\texttt{REGLUE} (800K steps, ImageNet 256{$\times$}256).} We vary the guidance interval $[0,\tau]$ and scale $w$. VAE-only applies CFG only to the VAE latents; otherwise CFG is applied to both VAE latents and VFM representations.}
    \vspace{-8pt}
    \begin{tabular}{lcccccccc}
\toprule
 $\Th{Interval}$  & $w$ & $\Th{VAE-Only}$ &  $\Th{FID}\downarrow$ & $\Th{sFID}\downarrow$ & $\Th{IS}\uparrow$ & $\Th{Prec.}\uparrow$ & $\Th{Rec.}\uparrow$ \\
\midrule
$[0, 0.85]$  & 2.8 & False &  1.55 & 4.30 & 278.20 & 0.77 & 0.66 \\
$[0, 0.90]$  & 2.8 & False &  1.53 & 4.26 & 320.43 & 0.78 & 0.65 \\
$[0, 0.95]$  & 2.8 & False &  2.75 & 4.20 & 395.91 & 0.82 & 0.60 \\

\arrayrulecolor{black!30}\cmidrule(lr){1-8}\arrayrulecolor{black}

$[0, 0.90]$  & 2.8 & False &  1.53 & 4.26 & 320.43 & 0.78 & 0.65 \\
$[0, 0.90]$  & 2.8 & True &  1.84 & 4.51 & 235.28 & 0.76 & 0.66 \\

\arrayrulecolor{black!30}\cmidrule(lr){1-8}\arrayrulecolor{black}

$[0, 0.90]$  & 2.8 & False &  1.53 & 4.26 & 320.43 & 0.78 & 0.65 \\
$[0, 0.90]$  & 2.7 & False &  1.52 & 4.29 & 315.02 & 0.78 & 0.65 \\
$[0, 0.90]$  & 2.6 & False &  1.49 & 4.27 & 310.09 & 0.78 & 0.65 \\
$[0, 0.90]$  & 2.5 & False &  1.48 & 4.26 & 305.14 & 0.78 & 0.65 \\
$[0, 0.90]$  & 2.4 & False &  1.47 & 4.24 & 299.74 & 0.78 & 0.65 \\
$[0, 0.90]$  & 2.3 & False &  1.46 & 4.23 & 293.85 & 0.78 & 0.65 \\
$[0, 0.90]$  & 2.2 & False &  1.53 & 4.26 & 320.43 & 0.78 & 0.65 \\

\arrayrulecolor{black!30}\cmidrule(lr){1-8}\arrayrulecolor{black}
$[0, 0.85]$  & 2.3 & False &  1.58 & 4.24 & 257.44 & 0.77 & 0.65 \\
$[0, 0.90]$  & 2.3 & False &  1.46 & 4.23 & 293.85 & 0.78 & 0.65 \\
$[0, 0.95]$  & 2.3 & False &  2.19 & 4.16 & 362.53 & 0.82 & 0.60 \\
\arrayrulecolor{black!30}\cmidrule(lr){1-8}\arrayrulecolor{black}

$[0, 0.90]$  & 2.3 & False &  1.46 & 4.23 & 293.85 & 0.78 & 0.65 \\
$[0, 0.90]$  & 2.3 & True &  1.76 & 4.39 & 240.73 & 0.76 & 0.66 \\
\bottomrule
\end{tabular}
    \label{tab:model_cfg_metrics}
\end{table}

\subsection{Small-scale \& high-resolution dataset.}
\label{sec:small_scale_hr_benchnmarks}
To test the robustness of our approach, we evaluated performance on the Food-101 ~\cite{food} (different domain $\&$ small-scale) and ImageNet-512 (high-resolution) datasets using \texttt{SiT-B/2}. As shown in~\autoref{tab:small_scale_hr_benchnmarks}, \ours consistently outperforms established baselines, achieving a significant lead in FID scores without the use of Classifier-Free Guidance (CFG).

\subsection{FID vs Training Compute}
\label{sec:fid_vs_training_compute}
We evaluate the training compute required by \ours relative to \reg at matched FID.
\ours processes the \emph{same} number of tokens as \reg: by design, local semantics
are fused \emph{channel-wise} with the VAE latents rather than concatenated along the
sequence, precisely to avoid any sequence-length overhead (see~\autoref{tab:fusion_strategy}).
See \autoref{tab:acceleration} for throughput (img/s) comparisons. Total overhead is
negligible, caused by our lightweight semantic compressor, yet \ours consistently
outperforms \reg at every matched compute point (\autoref{tab:flops_comparison}).
At XL scale, \ours matches \reg's 1M-step FID at only 700K steps---30\% less
compute---confirming that the gains are not a result of extra budget.

\begin{table}[t]
    \vspace{-11pt}
    \centering
    \scriptsize
    \setlength{\tabcolsep}{1.5pt}
    \caption{\textbf{FID vs.\ training compute} (ImageNet $256\times256$, no CFG). \ours vs.\ \reg
across training budgets; PFLOPs/run include \texttt{DINO-v2} + semantic compressor for \ours.
Subscripts show relative change vs.\ \reg.}
    \renewcommand{\arraystretch}{0.95}
    % shorthand for the tiny percentage subscript
    \newcommand{\pct}[2]{$_{\textcolor{#1}{\fontsize{3pt}{3pt}\selectfont(#2)}}$}
    \begin{tabular}{@{}llcc r@{}l@{}}
        \toprule
        \Th{Model} & \Th{Method} & \Th{Steps}
          & {Total train compute (\Th{PFLOPs/run})}
          & \multicolumn{2}{c}{\Th{FID$\downarrow$}} \\
        \midrule
        \multirow{3}{*}{\texttt{SiT-B/2}}
          & \reg  & 400K
            & 19{,}861\pct{gray}{0\%}\quad\quad\quad
            & 15.2     & \pct{gray}{0\%} \\
          & \ours & 300K
            & 15{,}286\pct{ForestGreen}{-23.0\%}
            & 14.5     & \pct{ForestGreen}{-4.6\%} \\
          & \ours & 400K
            & 20{,}381\pct{red}{+2.6\%}\enspace
            & \textbf{12.9} & \pct{ForestGreen}{-15.1\%} \\
        \midrule
        \multirow{3}{*}{\texttt{SiT-XL/2}}
          & \reg  & 1M
            & 197{,}493\pct{gray}{0\%}\quad\quad\quad
            & 2.7       & \pct{gray}{0\%} \\
          & \ours & 700K
            & 139{,}169\pct{ForestGreen}{-29.5\%}
            & 2.7 & \pct{gray}{0\%} \\
          & \ours & 1M
            & 198{,}812\pct{red}{+0.7\%}\enspace
            & \textbf{2.5} & \pct{ForestGreen}{-7.4\%} \\
        \bottomrule
    \end{tabular}

    \label{tab:flops_comparison}
\end{table}

\begin{table}[h]
    \centering
    \footnotesize
    \setlength{\tabcolsep}{2.1pt}
    \caption{\textbf{FID results with \texttt{SiT-B/2}} on Food \& ImageNet-512. Results without CFG.}
    \vspace{-8pt}
    \begin{tabular}{@{}lcccc@{}}
        \toprule
        \Th{Model} & \Th{Iter.} & \Th{Food-101} & \Th{ImageNet-512} \\
        \midrule
        \repa   & $400\text{K}$ & 14.5  & 25.7   \\
        %\redi    & 9.2   & -   & -    \\
        \reg    & $400\text{K}$ & 7.6 & 18.1 \\
        \rowcolor{TableColor}\ours & $400\text{K}$  & \textbf{4.6} & \textbf{14.4} \\
        \bottomrule
    \end{tabular}
\label{tab:small_scale_hr_benchnmarks}
\end{table}

\subsection{Fusion strategy.}
\label{sec:fusion_strategy}

We further ablate our fusion strategy by comparing the proposed \emph{channel-wise} fusion against a standard \emph{sequence-level} concatenation. In the concatenation setup, the input sequence length increases to 513 tokens, composed of 256 VAE latents, 256 patch-level semantic features, and a single global [CLS] token. 

As shown in~\autoref{tab:fusion_strategy}, sequence-level concatenation offers a negligible 4 \% FID gain but slashes throughput by 55\% (3.8 to 1.7 images/s) due to quadratic attention cost. We use channel-wise fusion by default for its vastly superior efficiency-to-quality ratio. Thus, channel-wise fusion offers the best trade-off and remains our default strategy.

\begin{table}[h]
 \centering
 \footnotesize
 \setlength{\tabcolsep}{2.1pt}
 \caption{\textbf{Fusion strategy ablation.} Results use \texttt{SiT-B/2}.}
 \vspace{-8pt}
\begin{tabular}{llll}
\toprule
\Th{Token Fusion Strategy} & \Th{\#Tokens} & \Th{Throughput (img/s) $\uparrow$}  & \Th{FID$\downarrow$} \\ 
\midrule
\rowcolor{TableColor} Channel-wise  & $257^{\:}$ & $3.8$  & $12.9$ \\
Sequence-level  &$513^\dagger$  & $1.7_{\textcolor{red}{\tiny(-55\%)}}$ & $12.3_{\textcolor{ForestGreen}{\tiny(-4\%)}}$ \\
\bottomrule
\end{tabular}
\label{tab:fusion_strategy}
\end{table}

\subsection{Resampling VFM grid}
\label{sec:resampling}
We evaluate the impact of interpolation kernels for resizing $\mathcal{E}_\psi(\rvf_\ast)$ to the $H_z \times W_z$ grid. As shown in~\autoref{tab:resampling}, bilinear resampling achieves the best performance, with negligible difference compared to the nearest alternative.

\begin{table}[h]
\centering
 \footnotesize
 \setlength{\tabcolsep}{2.1pt}
 \caption{Impact of resampling on image quality.}
 \vspace{-8pt}

\begin{tabular}{lc}
\toprule
Method & FID ($\downarrow$) \\ \midrule
Nearest Neighbor & 13.2 \\
\textbf{Bilinear} & \textbf{12.9} \\ \bottomrule
\end{tabular}
\label{tab:resampling}
\end{table}

\subsection{Limited data}

\begin{figure}[t]
\centering
\begin{tikzpicture}
\tikzset{
    gap label/.style={
        font=\scriptsize\bfseries,
        text=red!70!black,
        anchor=west,
        xshift=2pt,
        fill=white,
        fill opacity=0.85,
        text opacity=1,
        inner sep=1pt
    }
}

\begin{axis}[
  width=0.7\linewidth, height=0.67\linewidth,
  grid=both,
  grid style={dashed, gray!30},
  tick label style={font=\footnotesize},
  label style={font=\small},
  title style={font=\small},
  every axis plot/.append style={line width=1.1pt, mark options={solid}, mark size=1.5pt},
  xmin=18.0, xmax=146, 
  ymin=10, ymax=41.5,    
  xtick={20,50,100},
  xticklabels={20,50,100},
  scaled x ticks=false,
  xlabel={ImageNet (\%)}, ylabel={FID (↓)},
  xmode=log,           
  cycle list name=auxsched,
  legend columns=3,
  legend style={
    draw=none,
    font=\footnotesize,
    at={(0.5,-0.25)}, 
    anchor=north,
    /tikz/every even column/.append style={column sep=0.6em}}
]

% REG (baseline, keep color but dashed)
\addplot+[
  very thick,
  mark=*
] coordinates {(20,40.2) (50,21.0) (100,15.2)};
\addlegendentry{\reg}

% REGLUE (ours, keep color but solid + slightly emphasized)
\addplot+[
  very thick,
  mark=*
] coordinates {(20,34.7) (50,17.6) (100,12.9)};
\addlegendentry{\ours (ours)}

\addplot[
  only marks,
  mark=*,
  mark size=0.2pt,
  draw=red!70!black,
  fill=red!70!black
] coordinates {
  (20,37.5)   % (40.2 + 34.7)/2
  (50,19.3)   % (21.0 + 17.6)/2
  (100,14.05) % (15.2 + 12.9)/2
};

\addplot[
    only marks,
    mark=none,
    nodes near coords,
    point meta=explicit symbolic,
    every node near coord/.style={gap label}
] coordinates {
    (20.0,37.5)   [{$\Delta 5.5$}]
    (50.7,19.3)   [{$\Delta 3.4$}]
    (100,13.9) [{$\Delta 2.3$}]
};

\end{axis}
\end{tikzpicture}

\caption{\textbf{Dataset pruning on ImageNet.} 
FID on ImageNet $256{\times}256$ for \texttt{SiT-B/2} trained for $80$ epochs on class-balanced subsets (20\%, 50\%, 100\% of ImageNet). 
\ours consistently outperforms \reg, with improvements of $-5.5$, $-3.4$, and $-2.3$ FID at 20\%, 50\%, and 100\%, respectively.}
\label{fig:data_pruning}
\end{figure}
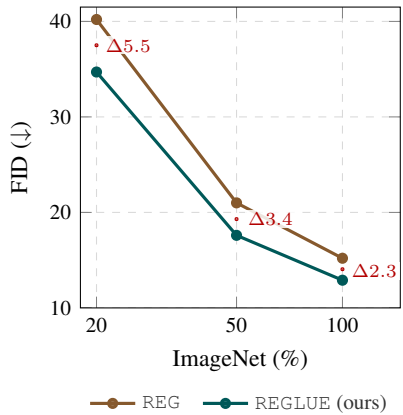

In~\autoref{fig:data_pruning}, we evaluate data efficiency by training \texttt{SiT-B/2} for 80 epochs on class-balanced ImageNet sets of 20\%, 50\%, and 100\%. \texttt{REGLUE} consistently outperforms \texttt{REG}, with larger gains when data is scarce: -5.5 FID at 20\% and -3.4 at 50\%. This indicates that jointly modeling compact local and global VFM semantics improves robustness on data-limited regimes.

\section{Additional Experimental Setup}

\subsection{Semantic compressor details}
\label{sec:compressor_details}
\noindent\textbf{Architecture settings.}
The compression model is a lightweight convolutional autoencoder composed of the \emph{semantic compressor}, which encodes the high-dimensional VFM features into a compact representation and the \emph{semantic de-compressor}, which symmetrically decodes them back to their original space. The detailed architecture is presented in~\autoref{fig:compressor_overview}. The semantic encoder is composed of three main components: an \emph{input layer}, a \emph{middle block}, and an \emph{output layer}. The input layer is a 3$\times$3 convolutional layer (3072$\rightarrow$256), where 3072 corresponds to the number of input channels from the concatenated VFM features and 256 denotes the hidden size. The middle block is a residual block (Conv–BN–ReLU–Conv–BN, 256 channels, identity skip) that preserves spatial shape. The output layer is a convolutional layer (256$\rightarrow$16) that projects the representation to 16 compressed channels. A symmetric semantic de-compressor mirrors this design (16$\rightarrow$256$\rightarrow$3072). The model is fully convolutional, preserves the spatial resolution, and is trained with an MSE reconstruction loss; at inference we retain only the encoder to provide compact local semantics.

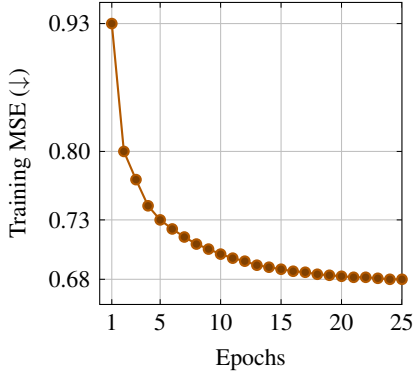
\begin{figure}[t]
\centering
\begin{tikzpicture}
\begin{axis}[
  width=5.6cm, height=5.6cm,
  xlabel={Epochs},
  ylabel={Training MSE (↓)},
  xmin=0, xmax=25,
  ymin=0.65, ymax=0.95,
  xtick={1,5,10,15,20,25},
  ytick={0.929,0.802,0.734,0.675},
  tick style={black},
  xticklabel style={/pgf/number format/fixed}, 
  yticklabel style={/pgf/number format/.cd, fixed, fixed zerofill, precision=2},
  tick label style={/pgf/number format/fixed},
  major tick length=2pt,
  axis line style={black},
  enlargelimits=false
]

\addplot+[color=orange!70!black,
  mark=*,
  mark options={draw=orange!70!black, fill=orange!50!black}, 
  thick]
  coordinates {(1, 0.929) (2, 0.802) (3, 0.774) (4,0.748) (5,0.734) (6,0.725) (7,0.717) (8,0.71) (9,0.705) (10,0.7) (11,0.696) (12,0.693) (13,0.689) (14,0.687) (15,0.685) (16,0.683) (17,0.682) (18,0.68) (19,0.679) (20,0.678) (21,0.677) (22,0.677) (23,0.676) (24,0.675) (25,0.675) };

\end{axis}
\end{tikzpicture}
\vspace{-8pt}
\caption{\textbf{Compressor training.} Training curve showing MSE loss over epochs. The compression model utilizes 4 last \texttt{DINOv2-B} layers. The Input layer is 256 and the compression is to 16 channels. A run finishes in \emph{less than one hour} on \(8\times\)A100 GPUs.}
\label{fig:compressor_training}
\end{figure}

\vspace{4pt}
\noindent\textbf{Optimization settings.}
We train the semantic compressor for 25 epochs with an MSE reconstruction loss between the concatenated multi-layer VFM features and their decoded counterparts. We use Adam~\cite{kingma2017adammethodstochasticoptimization} with a learning rate of \(1\times 10^{-3}\), $(\beta_1, \beta_2)=(0.9, 0.999)$, batch size 4096, and no weight decay. The learning rate decays with a cosine schedule to a final value of \(8.5\times 10^{-4}\). \autoref{fig:compressor_training} plots the training curve: the loss decreases smoothly and plateaus by the final epoch, indicating stable convergence. The model is lightweight; a full run finishes in \emph{under one hour} on \(8\times\)A100 GPUs.

\subsection{SiT details}
\label{sec:sit_details}

\noindent\textbf{Architecture settings.} We adopt the official \texttt{SiT} configurations~\cite{ma2024sit}. The base \texttt{SiT-B/2} (132M params) uses 12 transformer blocks with embedding dimension \(768\) and 12 attention heads. The larger \texttt{SiT-XL/2} (677M params) uses 28 blocks with embedding dimension \(1152\) and 16 heads.

\begin{table}[h!]
    \centering
    \small
    \setlength{\tabcolsep}{3pt}
    \caption{\textbf{\texttt{SiT} optimization settings.}}
    \vspace{-8pt}
    \begin{tabular}{lcccc}
        \toprule
        \mc{2}{\Th{Optimization}} \\
        \midrule
        Batch size  & 256 \\ 
        Optimizer  & AdamW \\
        Learning Rate  & 0.0001 \\
        $(\beta_1, \beta_2)$  & (0.9, 0.999) \\
        \midrule
        \mc{2}{\Th{Interpolants}} \\
        \midrule
        $\alpha_t$  & $1-t$ \\
        $\sigma_t$  & $t$  \\
        Training objective  & v-prediction\\
        Sampler  & Euler-Maruyama  \\
        Sampling steps  & 250 \\
        \bottomrule
    \end{tabular}
    \label{tab:hyperparam_opt}
\end{table}

\vspace{4pt}
\noindent\textbf{Optimization settings.} We use AdamW~\cite{kingma2017adammethodstochasticoptimization, loshchilov2019decoupledweightdecayregularization} with a constant learning rate of $1 \times 10^{-4}$, $(\beta_1, \beta_2)=(0.9, 0.999)$, and a batch size of 256 for both \texttt{SiT} models. To speed up training, we use mixed-precision ($\mathrm{fp16}$) with gradient clipping. We also pre-compute image latent using $\mathrm{SD-VAE}$ \cite{rombach2022high}. The training objective is $\mathrm{v-prediction}$, and we use the Euler–Maruyama sampler with 250 steps, defining the interpolants as $\alpha_t = 1-t$ and $\sigma_t = t$. We provide the optimization details in~\autoref{tab:hyperparam_opt}.

\section{Limitations and Future Work}

As shown in~\autoref{tab:acceleration} and \autoref{tab:fid_comparison} in the main paper, \ours consistently improves sample quality under comparable training budgets and reaches or surpasses strong baselines in substantially fewer iterations. However, due to resource constraints, we restrict \texttt{SiT-XL/2}{+}\ours experiments to 1M iterations and do not explore full convergence at ultra–long schedules (\eg, 4M iterations). On our available compute (8$\times$A100 GPUs), a single 1M-iteration \texttt{SiT-XL/2} run requires roughly 7 days, making exploration of such long schedules impractical. Large scale higher-resolution ImageNet 512$\times$512 experiments (\eg, using \texttt{SiT-XL/2})  are an interesting next step as well; in this work, we instead prioritize configurations that we consider also interesting and practical, such as limited-data regimes (see~\autoref{fig:data_pruning}).   

Beyond scaling, our results point to several promising extensions. First, swapping \texttt{DINOv2} for \texttt{DINOv3} yields further gains (\autoref{tab:different_vfm} and \autoref{tab:ablationinit} in the main paper), suggesting that stronger VFMs could enhance \ours. Second, while we currently include the raw global \texttt{[CLS]} token, learning a compact \emph{global} compressor (analogous to our spatial one) may better balance global–local capacity within our joint modeling framework. Third, jointly decoding all latents (semantic + VAE) rather than only the VAE latents is an appealing extension; our preliminary trials showed mixed results across model scales, suggesting that a decoder design which robustly exploits the full latent set is worth further study.

\section{Baseline Generative Models}

We briefly summarize the baselines used in our comparisons. Autoregressive baselines include \texttt{VAR}~\cite{tian2024visual}, which progressively predicts fine image details from coarse inputs across multiple scales; \texttt{MagViTv2}~\cite{yu2024language}, which removes lookup tables in quantization to support much larger token vocabularies; and \texttt{MAR}~\cite{li2024autoregressive} which avoids vector quantization altogether in an autoregressive setup. 

For latent diffusion models, we consider \texttt{LDM}~\cite{rombach2022high}, which performs diffusion in a compact latent space; \texttt{DiT}~\cite{peebles2023scalable}, a transformer-based architecture; \texttt{U-ViT-H/2}~\cite{bao2023all} a ViT-based diffusion model with skip connections; \texttt{MaskDiT}~\cite{zheng2023fast}, which adds a mask-reconstruction auxiliary task; \texttt{MDT}~\cite{gao2023mdtv2}, which employs an asymmetric masked latent modeling scheme; \texttt{SD-DiT}~\cite{zhu2024sd}, which augments \texttt{MaskDiT} with a momentum-encoder–based discrimination loss; \texttt{SiT}~\cite{ma2024sit}, which recasts the \texttt{DiT} backbone within a continuous-time interpolant framework; and \texttt{FasterDiT}~\cite{yao2024fasterdit}, which accelerates training through velocity-supervised objectives. 

Finally, among methods that explicitly use visual representations, we include \texttt{REPA}~\cite{Yu2025repa}, which aligns diffusion-model features with local VFM features; \texttt{ReDi}~\cite{kouzelis2025redi} which linearly compress VFM features and directly model them in the diffusion model; and finally \texttt{REG}~\cite{wu2025representation} which models the global VFM representation while also applying \texttt{REPA}-style alignment.

\section{Visualizations}
\label{sec:visualizations}

\begin{figure*}[t!]
    \vspace{-15pt}
    \centering
{
\small
\centering
\newcommand{\resultsfignew}[1]{\includegraphics[width=0.09\textwidth,valign=t]{#1}}

\newcommand{\resultsfignewmedium}[1]{\includegraphics[width=0.184\textwidth,valign=t]{#1}}

\newcommand{\resultsfignewbig}[1]{\includegraphics[width=0.3755\textwidth,valign=t]{#1}}
\setlength{\tabcolsep}{1pt}

\setlength{\tabcolsep}{1pt}

\begin{tabular}{@{}ccccccc@{}}  % 7 columns total

\mc{7}{Class label = "Golden Retriever" (207)}\\

\multicolumn{4}{c}{%
    \multirow{4}{*}{%
       \resultsfignewbig{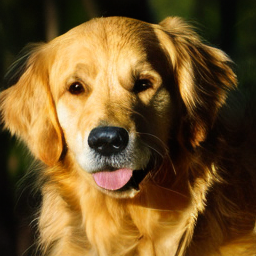}%
    }%
} & 

\multicolumn{2}{c}{%
    \multirow{2}{*}{%
       \resultsfignewmedium{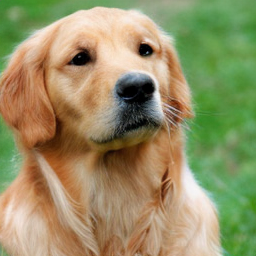}%
    }%
}
&
\resultsfignew{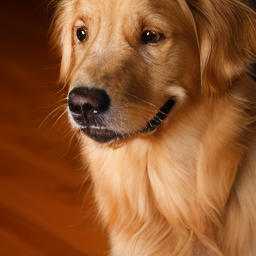}
\\
\mc{7}{\vspace{-2.0ex}}\\

& & & & 
& & 
\resultsfignew{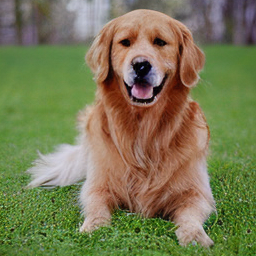} \\
\mc{7}{\vspace{-2.0ex}}\\

& & & & 
\multicolumn{2}{c}{%
    \multirow{2}{*}{%
       \resultsfignewmedium{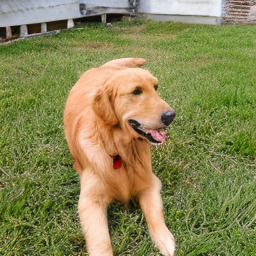}%
    }%
} & 
\resultsfignew{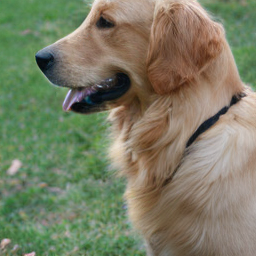} \\
\mc{7}{\vspace{-2.0ex}}\\

& & & & 
& & 
\resultsfignew{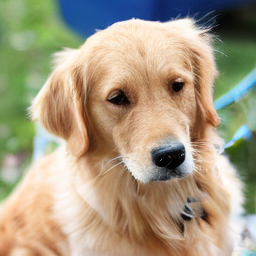} \\

\mc{7}{\vspace{-1.ex}}\\

\mc{7}{Class label = “Castle” (483)}\\

\multicolumn{4}{c}{%
    \multirow{4}{*}{%
       \resultsfignewbig{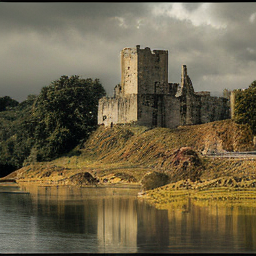}%
    }%
} & 

\multicolumn{2}{c}{%
    \multirow{2}{*}{%
       \resultsfignewmedium{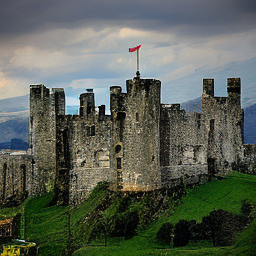}%
    }%
}
&
\resultsfignew{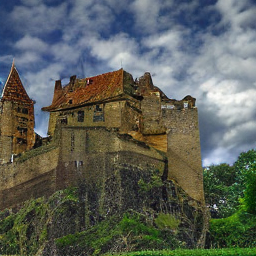}
\\
\mc{7}{\vspace{-2.0ex}}\\

& & & & 
& & 
\resultsfignew{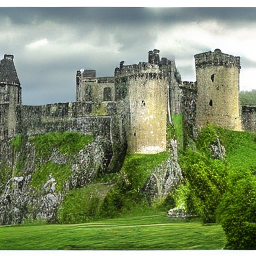} \\
\mc{7}{\vspace{-2.0ex}}\\

& & & & 
\multicolumn{2}{c}{%
    \multirow{2}{*}{%
       \resultsfignewmedium{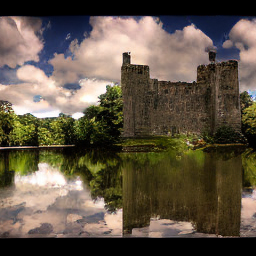}%
    }%
} & 
\resultsfignew{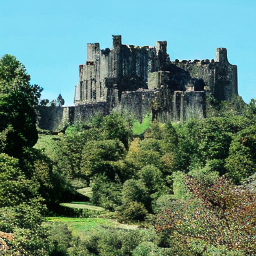} \\
\mc{7}{\vspace{-2.0ex}}\\

& & & & 
& & 
\resultsfignew{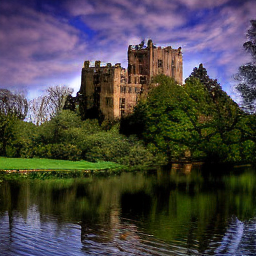} \\

\mc{7}{\vspace{-1.ex}}\\

\mc{7}{Class label = “Bald Eagle” (22)}\\

\multicolumn{4}{c}{%
    \multirow{4}{*}{%
       \resultsfignewbig{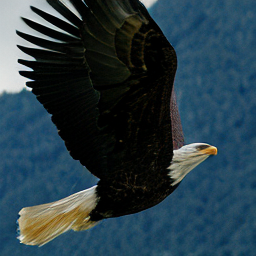}%
    }%
} & 

\multicolumn{2}{c}{%
    \multirow{2}{*}{%
       \resultsfignewmedium{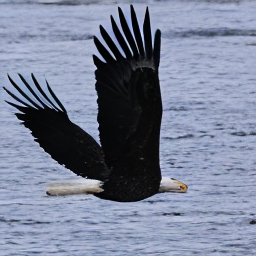}%
    }%
}
&
\resultsfignew{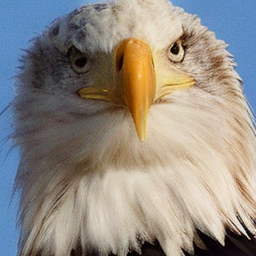}
\\
\mc{7}{\vspace{-2.0ex}}\\

& & & & 
& & 
\resultsfignew{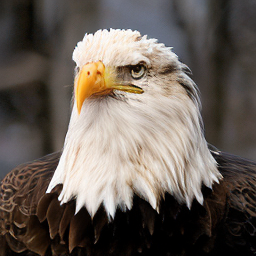} \\
\mc{7}{\vspace{-2.0ex}}\\

& & & & 
\multicolumn{2}{c}{%
    \multirow{2}{*}{%
       \resultsfignewmedium{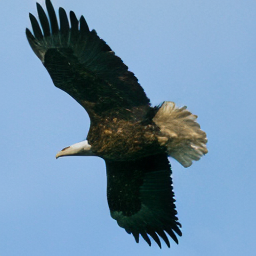}%
    }%
} & 
\resultsfignew{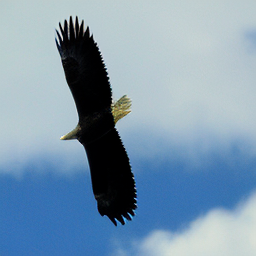} \\
\mc{7}{\vspace{-2.0ex}}\\

& & & & 
& & 
\resultsfignew{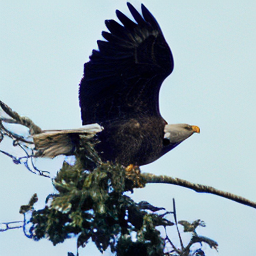} \\
    
\end{tabular}
}
    \vspace{-2pt}
    \caption{\textbf{Uncurated ImageNet $\mathbf{256 \times 256}$ samples.} Class-conditional generations from \texttt{SiT-XL/2+}\ours trained for 1M steps with CFG ($w{=}4.0$). Grids illustrate great fidelity and intra-class diversity.}
    \label{fig:appendix-gen-1}
    \vspace{-10pt}
\end{figure*}

\begin{figure*}[t!]
    \vspace{-15pt}
    \centering
{
\small
\centering
\newcommand{\resultsfignew}[1]{\includegraphics[width=0.09\textwidth,valign=t]{#1}}

\newcommand{\resultsfignewmedium}[1]{\includegraphics[width=0.184\textwidth,valign=t]{#1}}

\newcommand{\resultsfignewbig}[1]{\includegraphics[width=0.3755\textwidth,valign=t]{#1}}
\setlength{\tabcolsep}{1pt}

\setlength{\tabcolsep}{1pt}

\begin{tabular}{@{}ccccccc@{}}  % 7 columns total

\mc{7}{Class label = "Bee" (309)}\\

\multicolumn{4}{c}{%
    \multirow{4}{*}{%
       \resultsfignewbig{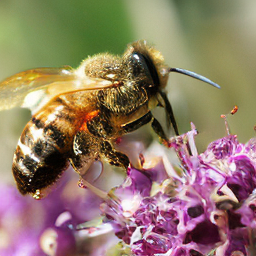}%
    }%
} & 

\multicolumn{2}{c}{%
    \multirow{2}{*}{%
       \resultsfignewmedium{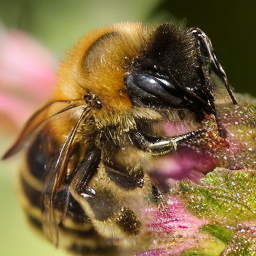}%
    }%
}
&
\resultsfignew{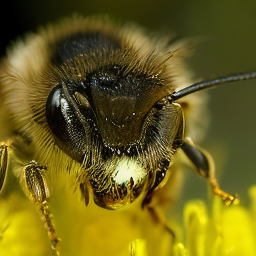}
\\
\mc{7}{\vspace{-2.0ex}}\\

& & & & 
& & 
\resultsfignew{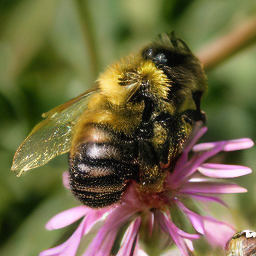} \\
\mc{7}{\vspace{-2.0ex}}\\

& & & & 
\multicolumn{2}{c}{%
    \multirow{2}{*}{%
       \resultsfignewmedium{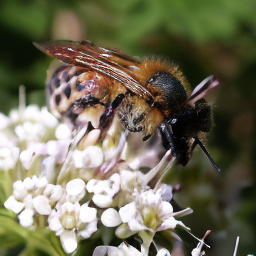}%
    }%
} & 
\resultsfignew{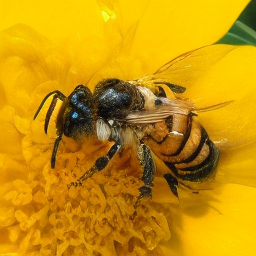} \\
\mc{7}{\vspace{-2.0ex}}\\

& & & & 
& & 
\resultsfignew{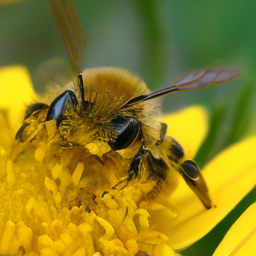} \\

\mc{7}{\vspace{-1.ex}}\\

\mc{7}{Class label = “Great Grey Owl” (24)}\\

\multicolumn{4}{c}{%
    \multirow{4}{*}{%
       \resultsfignewbig{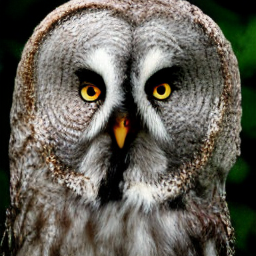}%
    }%
} & 

\multicolumn{2}{c}{%
    \multirow{2}{*}{%
       \resultsfignewmedium{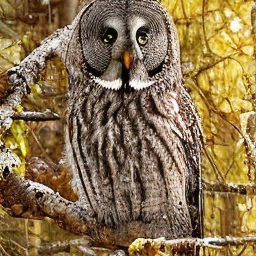}%
    }%
}
&
\resultsfignew{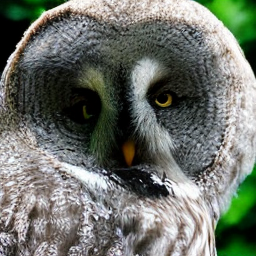}
\\
\mc{7}{\vspace{-2.0ex}}\\

& & & & 
& & 
\resultsfignew{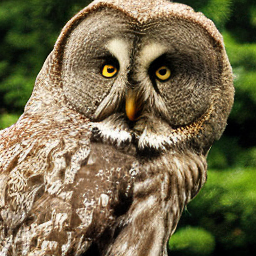} \\
\mc{7}{\vspace{-2.0ex}}\\

& & & & 
\multicolumn{2}{c}{%
    \multirow{2}{*}{%
       \resultsfignewmedium{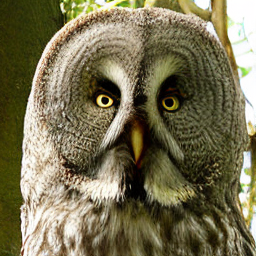}%
    }%
} & 
\resultsfignew{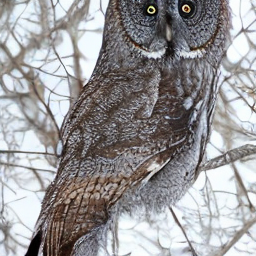} \\
\mc{7}{\vspace{-2.0ex}}\\

& & & & 
& & 
\resultsfignew{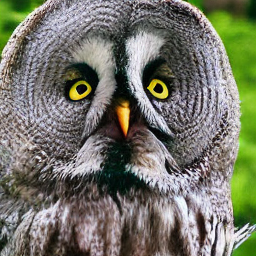} \\

\mc{7}{\vspace{-1.ex}}\\

\mc{7}{Class label = “Cheeseburger” (933)}\\

\multicolumn{4}{c}{%
    \multirow{4}{*}{%
       \resultsfignewbig{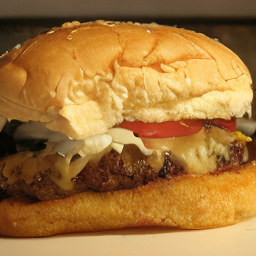}%
    }%
} & 

\multicolumn{2}{c}{%
    \multirow{2}{*}{%
       \resultsfignewmedium{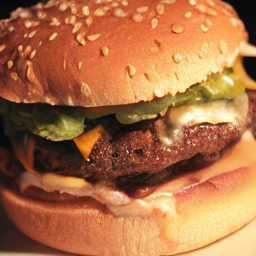}%
    }%
}
&
\resultsfignew{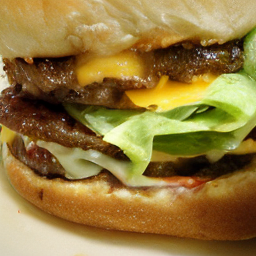}
\\
\mc{7}{\vspace{-2.0ex}}\\

& & & & 
& & 
\resultsfignew{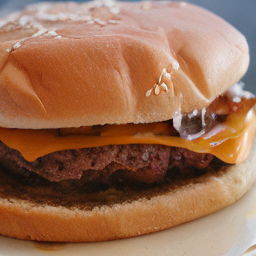} \\
\mc{7}{\vspace{-2.0ex}}\\

& & & & 
\multicolumn{2}{c}{%
    \multirow{2}{*}{%
       \resultsfignewmedium{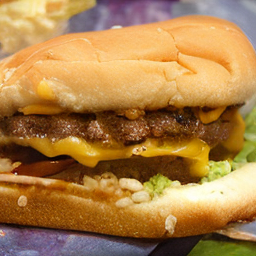}%
    }%
} & 
\resultsfignew{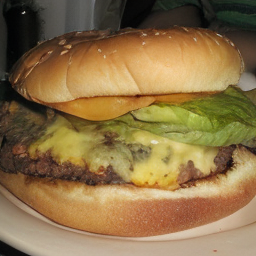} \\
\mc{7}{\vspace{-2.0ex}}\\

& & & & 
& & 
\resultsfignew{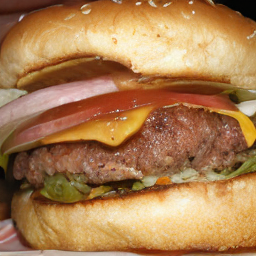} \\
    
\end{tabular}
}
    \vspace{-2pt}
    \caption{\textbf{Uncurated ImageNet $\mathbf{256 \times 256}$ samples.} Class-conditional generations from \texttt{SiT-XL/2+}\ours trained for 1M steps with CFG ($w{=}4.0$). Grids illustrate great fidelity and intra-class diversity.}
    \label{fig:appendix-gen-2}
    \vspace{-25pt}
\end{figure*}

We present uncurated, class-conditional samples from \texttt{SiT-XL/2+}\ours trained for 1M steps at $256{\times}256$ in~\autoref{fig:appendix-gen-1} and~\autoref{fig:appendix-gen-2}. We use CFG  with $w{=}4.0$. The grids illustrate both fine-grained detail (textures and object parts) and diversity within each class.

\clearpage

\end{document}